\documentclass[twoside,11pt]{article}
\usepackage{jmlr2e}

\usepackage{url}
\usepackage{natbib}
\usepackage{graphicx}
\usepackage{amsmath}
\usepackage{amsfonts}

\ShortHeadings{Segregating event streams}{Stowell and Plumbley}
\firstpageno{1}

\begin{document}

\title{Segregating event streams and noise \\with a Markov renewal process model}

%\author{Dan Stowell and Mark D. Plumbley\\%
%Centre for Digital Music, Queen Mary, University of London\\
%\texttt{dan.stowell@eecs.qmul.ac.uk}}
\author{\name Dan Stowell \email dan.stowell@eecs.qmul.ac.uk\\
\name Mark D. Plumbley \email mark.plumbley@eecs.qmul.ac.uk\\
\addr Centre for Digital Music\\
Queen Mary University of Londdon}

\editor{Editor name}

\maketitle

%%%%%%%%%%%%%%%%%%%%%%%%%
\begin{abstract}%
We describe an inference task in which a set of timestamped event observations
must be clustered into an unknown number of temporal sequences
with independent and varying rates of observations.
Various existing approaches to multi-object tracking assume a fixed number of sources 
and/or
a fixed observation rate;
we develop an approach to inferring structure in timestamped data produced by a mixture
of an unknown and varying number of similar Markov renewal processes, plus independent clutter noise. 
The inference simultaneously distinguishes signal from
noise as well as clustering signal observations into separate source streams.
We illustrate the technique via a synthetic experiment 
as well as an experiment to track a mixture of singing birds. 
\end{abstract}


%%%%%%%%%%%%%%%%%%%%%%%%%
\section{Introduction}
\label{sec:intro}

Various approaches exist for the task of inferring the temporal evolution of multiple sources based on joint observations \citep{Mahler:2007a,VanGael:2008}.
They are generally based on
a model in which sources are continuously observable, in the sense that they are expected to emit/return observations at every time step (though there may be missed detections).
Yet there are various types of source for which observations are inherently intermittent, and for which this intermittence exhibits temporal structure
that can be characterised as a point process.
Examples include 
sound event sequences such as bird calls or footsteps \citep{Wang:2006},
internet access logs \citep{Arlitt:1997}, 
pulsars in astronomy \citep{Keane:2010}
and neural firing patterns \citep{Bobrowski:2009}.
Intermittent observations are also often output from \textit{sparse representation} techniques, which transform signals into a representation with activations distributed sparsely in time and state \citep{Plumbley:2010}.

In this paper we describe a generic problem setting that may be applied to such data, along with an approach to estimation.
We are given a set of timestamped data,
and we assume each datum is produced by one of a set of similar but independent signal processes, or by a ``clutter'' noise process, with known parameters.
We do not know the true partitioning of the data into sequences each generated by a single process, and wish to infer this.
We do not know how many processes are active, and we do not assume that each process produces the same number of observations, or observations at the same time points.

This specific type of clustering problem has applications in various domains.
For example, when sparse representation techniques are used for source separation in time series, 
they often yield a set of atomic activations which must be clustered according to their underlying source, and preferably to discard any spurious noise activations \citep{Plumbley:2010}.
Temporal dependence information may help to achieve this (cf. \citet{Mysore:2010}).
Timestamped data such as internet access logs often contain no explicit user association,
yet it may be desirable to group such data by user for for further analysis \citep{Arlitt:1997}.
In computational audio scene analysis,
it is often the case that sound sources emit sound only intermittently during their presence in the scene (e.g.\ bird calls, footsteps),
yet it is desirable to track their temporal evolution \citep{Wang:2006}.

\subsection{Related Work}

To our knowledge, this particular problem setting has not been directly addressed in the literature.
Temporal data is most commonly treated using a model of sources which update continuously, or synchronously at an underlying temporal sampling rate.
Pertinent formulations for our purposes include the infinite factorial hidden Markov model (infinite FHMM) of \citet{VanGael:2008},
or the probability hypothesis density filter (PHD filter) \citep{Mahler:2007a},
both of which infer an unknown number of independent Markov sources.
FHMMs assume that the underlying sources are not intermittent during their lifetime,
and also that they persist throughout the whole observation period.
Pragmatically, intermittent emissions may be handled by incorporating silence states,
though the duration of such states cannot take an arbitrary distribution.
The PHD filter allows for stochastic missed detections but not for structured intermittency.

Among techniques which do not assume a synchronous update,
graph clustering approaches such as normalised cuts have similarities to our approach \citep{Shi:2000}.
In particular, \citet{Lagrange:2008} apply normalized cuts in order to cluster temporally-ordered data.
However, the normalised cuts method is applied to undirected graphs, 
and \citet{Lagrange:2008} use perceptually-motivated similarity criteria rather than directed Markov dependencies as considered herein.
Further, the normalized cuts method does not include a representation of clutter noise, 
and so \citet{Lagrange:2008} perform signal/noise cluster selection as a separate postprocessing step.
In the present work we include an explicit noise model.

Our problem setting also exhibits similarities with that of structure discovery in Bayesian networks \citep{Koivisto:2004}.
However, in that context the dependency structure is inferred from correlations present in multiple observations from each vertex in the structure.
In the present case we have only one observation per vertex,
plus the partial ordering implied by temporality.

In the following
we develop a model in which an unknown number of point-process sources are assumed to be active as well as Poisson clutter,
and describe how to perform a maximum likelihood inference which clusters the signal into individual identified tracks plus clutter noise.
We then demonstrate the performance of the approach in synthetic experiments,
and in an experiment analysing birdsong audio.

%%%%%%%%%%%%%%%%%%%%%%%%%
\section{Preliminaries}
\label{sec:prelim}

Throughout we will consider sets of observations in the form $\{(X, T)\}$ where $X$ is state and $T$ is time.
A Markov renewal process (MRP) generates a sequence of such observations
having the Markov property:
\begin{align}
      P(\tau_{n+1} \le t, X_{n+1}=j|(X_1, T_1),\ldots, (X_n=i, T_n)) & \nonumber \\
   =  P(\tau_{n+1} \le t, X_{n+1}=j|X_n=i) & %\nonumber \\
                   %& 
		   \qquad \qquad
		   \, \forall n \ge1, \,  t\ge0, \,   i,j \in \mathcal{S}
\end{align}
where $\tau_{n+1}$ is the time difference $T_{n+1} - T_n$.
Note that $\tau$ is not explicitly given in observations $\{(X,T)\}$, 
but can be inferred if we know that a particular pair of observations
are adjacent members within a sequence.

We will have cause to represent our data as a \textit{network flow} problem 
\citep[Chapter 3]{Bang-Jensen:2009}.
A \textit{network} is a graph supplemented such that each arc $A_{ij}$
has a \textit{lower capacity} $l_{ij}$ and \textit{upper capacity} $u_{ij}$, and a \textit{cost} $a_{ij}$.
A \textit{flow} is a function $x : A \rightarrow \mathcal{R}_0 $ that associates a value with each arc in the network.
We will be concerned with integer flows $x : A \rightarrow \mathcal{Z}_0 $.
A flow is \textit{feasible} if $l_{ij} \leq x_{ij} \leq u_{ij}$ for all $A_{ij}$ in the graph,
and for all vertices (except for any source/sink vertices) the sum of the inward flow is equal to the sum of the outward flow.
For any flow we can calculate a total cost as the sum of $a_{ij}x_{ij}$ over all $A_{ij}$.
We define the \textit{value} of a feasible flow to be the sum of $x_{ij}$ over all arcs leading from source vertices.

The standard terminology of flow networks associates capacities, flows and costs with arcs but not vertices.
However, in the following we will have cause to associate such attributes with vertices as well as with arcs.
This can be implemented transparently by the standard technique of \textit{vertex expansion},
in which each vertex is replaced by an in-vertex and an out-vertex,
plus a single arc between them which bears the associated attributes
\citep[Section 3.2.4]{Bang-Jensen:2009}.

%%%%%%%%%%%%%%%%%%%%%%%%%
\section{Mixtures of Markov Renewal Processes with Clutter Noise}
\label{sec:model}
% DESCRIBE MRP; MULTI-MRPS WITH BIRTH AND DEATH; PLUS CLUTTER

For the present task, we consider MRPs which are time-limited:
each process comes into being at a particular point in time
(governed by an independent Poisson process with intensity $\lambda_b(X)$),
and after each observation it may ``die'' with an independent death probability $p_d(X)$.
Otherwise
it transitions to a new random state-and-time according to the transition distribution $f_x(X, \tau)$.
The overall system to be considered is not one but a set of such time-limited MRPs,
plus a separate Poisson process that generates clutter noise with intensity $\lambda_c(X)$.
The MRPs are independent but share common parameters. % as given above.
We will refer to the overall system (including the noise process) as a \textit{multiple Markov renewal process} system or \textit{MMRP},
in order to clarify when we are referring to the whole system or to a single MRP.

We receive a set of $N$ observations in the form $\{(X, T)\}$ 
and we assume that they were generated by an MMRP
for which the process parameters are known,
but the number $K$ of MRPs is unknown
as well as the allocation of each observation to its generating process.
We assume that each observation is generated either by one MRP or by the noise process.
Given these observations
as well as model parameters $f_x(X, \tau)$, $\lambda_b$, $p_d$, $\lambda_c$,
there are many ways to cluster the observations into $K \in [0,N]$ non-overlapping subsets 
to represent the assertion that each cluster represents all the emissions from a single MRP,
with $H$ of the observations not included in any cluster and considered to be noise.
The overall likelihood under a chosen clustering is given by
\begin{align}
	\textrm{likelihood} = %
	\prod_{   k=1}^K{p_{\textrm{MRP}}(k)}  %
	\prod_{\eta=1}^H{p_{\textrm{NOISE}}(\eta)}  \nonumber
\end{align}
where
$p_{\textrm{MRP}}(k)$
represents the likelihood of the observation subsequence in cluster $k$ being generated by a single MRP,
and
$p_{\textrm{NOISE}}(\eta)$
represents the likelihood of a single observation datum under the noise model.
(A set of clusters is arbitrarily indexed by $k \in [1,K]$.)

In order to find the maximum likelihood solution,
we may equivalently divide the likelihood expression through by a constant factor, to give an alternative expression to be maximised.
We divide by the likelihood that all data were generated by the noise process, to give the likelihood ratio:
\begin{align}
	\textrm{L} =	\prod_{k=1}^K{  \frac{p_{\textrm{MRP}}(k)}{p_{\textrm{NOISE}}(k)} }
	\label{eqn:or}
\end{align}
where for notational simplicity we use 
$p_{\textrm{NOISE}}(k)$
as the joint likelihood of all observations contained within cluster $k$ under the noise model.
This likelihood ratio $L$ will shortly be seen to be a convenient expression to optimise.

The component likelihood ratio for a single cluster $k$
is given by

\begin{align}
	\frac{p_{\textrm{MRP}}(k)}{p_{\textrm{NOISE}}(k)}  % \nonumber \\ 
	=
	\frac{
	p_b(X_{k,1}) \, \cdot p_d(X_{k,n}) \, \cdot \prod_{i=2}^{n_k}{ f_{X_{k,i-1}}(X_{k,i}, T_{k,i} - T_{k,i-1}) }
		}{
		\prod_{i=1}^{n_k}{ p_c(X_{k,i})}
		}
		\label{eqn:ork}
\end{align}
where $(X_{k,i}, T_{k,i})$ refers to the $i$th observation assigned to cluster $k$, 
this cluster having $n_k$ observations indexed in ascending time order.
$p_d(\cdot)$ refers to the likelihood associated with a single observation under the Poisson process
parametrised by $\lambda_d$, 
and similarly for $p_c(\cdot)$ for the clutter process parametrised by $\lambda_c$.

The overall likelihood ratio $L$ tells us the relative likelihood that the observation set was generated by the selected clustering of signals and noise,
as opposed to the possibility that all observations were generated by clutter noise.
%Note that those observations not included in any of the $K$ clusters are attributed to the clutter noise process
%in both situations (namely, the situation that the clusters were generated by MRPs and the situation that the clusters were generated by the noise process).
%Hence their contribution to the likelihood ratio \eqref{eqn:or} is the same in the numerator and the denominator,
%and they can be omitted from the overall likelihood ratio calculation.
Our goal is to find the clustering that yields the highest likelihood ratio,
and therefore the set of MRP track identities that is most likely to originate from signal rather than noise.

\subsection{Network Flow Representation}
\label{sec:flow}

For any observation set of non-trivial size,
there is a combinatorial explosion of possible clusterings available
and enumerating them all is intractable.
In this subsection we propose to transform the problem into an equivalent problem of network flow,
which can be addressed using graph theoretic techniques.

To maximise the likelihood ratio, we can equivalently minimise its negative logarithm,
which we will consider as a ``cost'' for any particular solution.
We define additive component costs for birth, death, transition and clutter respectively as:
\begin{subequations}
\begin{align}
	a_b(X) &= -\log{ p_b(X) } 
	\\
	a_d(X) &= -\log{ p_d(X) } 
	\\
	a_t(X, X', \tau) &= -\log{ f_X(X', \tau) } 
	\\
	a_c(X) &= \log{ p_c(X) }
\end{align}
\label{eqn:costs}
\end{subequations}%
\noindent%
which leads to the following expression for the overall cost under a particular cluster assignment:
\begin{align}
	-\log(\textrm{L}) 
		= \sum_{k=1}^K & \left( 
		a_b(X_{k,1})
		+ a_d(X_{k,n})  \frac{}{} \right. 
			\nonumber \\
			& 
			+ \sum_{i=2}^{n_k}{ a_t(X_{ik,i-1}, X_{k,i}, T_{k,i} - T_{k,i-1} ) }
			\nonumber \\
			&   \left. 
		+ \sum_{i=1}^{n_k}{ a_c(X_{k,i})} 
		\right) .
		\label{eqn:logor}
\end{align}

\begin{figure}[t]
	\centering
	\includegraphics [width=0.7\textwidth]  {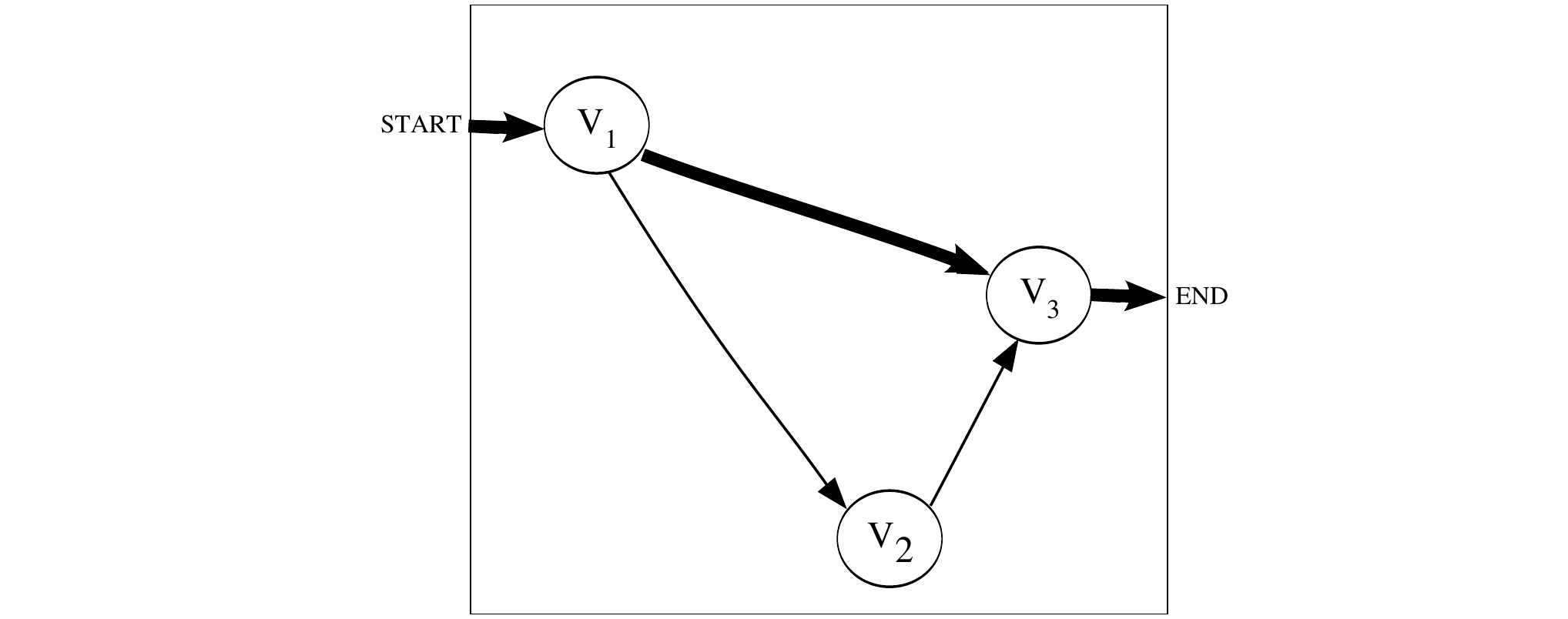} \\
	\caption{Simple illustration of a path within a network that might correspond to a single MRP sequence. %
	Time increases along the horizontal axis. %
	The bold arrows indicate a path from the first to the third datum %
	(the second datum being left out of the corresponding cluster). %
	The thin arrows indicate an alternative possible path. %
	}
\label{fig:ftransnet}
\end{figure}

The Markov structure of transitions,
as well as this representation as additive costs,
permit a natural representation as a problem defined on a directed graph.
If we construct a directed graph with observations as vertices and possible transitions as arcs,
then every possible path in the graph
(from any vertex to any other reachable vertex)
corresponds to one potential MRP cluster
(Figure \ref{fig:ftransnet}).
A set of $K$ paths corresponds to a set of $K$ MRP clusters.
To reflect the assumption that each observation is generated by no more than one MRP,
we require that a vertex can be a member of no more than one path in such a set.
Vertices not included in any of the paths correspond to noise observations.
%Note however that we permit singleton paths (containing only one node),
%%which would correspond to a birth and immediate death without transition;
%these would be labelled as signal rather than noise,
%if their likelihood outweighed the corresponding noise likelihood.

Given our restriction that a vertex can be included in no more than one path,
the problem of finding a mutually compatible set of MRP clusterings is equivalent to solving a particular kind of \textit{network flow} problem %in the graph we have constructed 
\citep[Chapter 3]{Bang-Jensen:2009}.
In our case, the concept of a flow will be used to pick out a set of arcs in the graph corresponding to a possible clustering,
by associating each arc with a value 1 or 0 indicating whether the arc is included in the clustering.
Therefore, in addition to the requirement that the flow is integer-valued,
all arcs will be defined to have unit capacity:
$l_{ij} = 0, u_{ij} = 1$ for all $A_{ij}$.
To reflect our assumption that each observation can be included in only one cluster, 
we will also specify unit capacities for all vertices. % except the source and sink;

It remains to specify how we can associate the costs \eqref{eqn:costs}
with the network such that we can solve for the minimum-cost solution to \eqref{eqn:logor}.
Transition costs will be associated with arcs,
and clutter costs with vertices,
but in order to include birth and death costs
we must modify the network by adding a single ``source'' vertex with an outward arc to all other vertices,
and a single ``sink'' vertex with an inward arc from all other vertices,
and by requiring that no other vertices act as sources or sinks (i.e.\ in a feasible flow, their inward and outward flows must balance).
We then associate birth costs with arcs from the source and death costs with arcs to the sink.
This means that all feasible flows in our network will be composed of paths
which consist of one single birth cost,
plus a sequence of clutter and transition costs,
and a single death cost.
The source and sink have infinite capacity,
allowing for solutions with unbounded $K$.

\begin{figure*}[t]
	\centering
	\includegraphics [width=0.7\textwidth]  {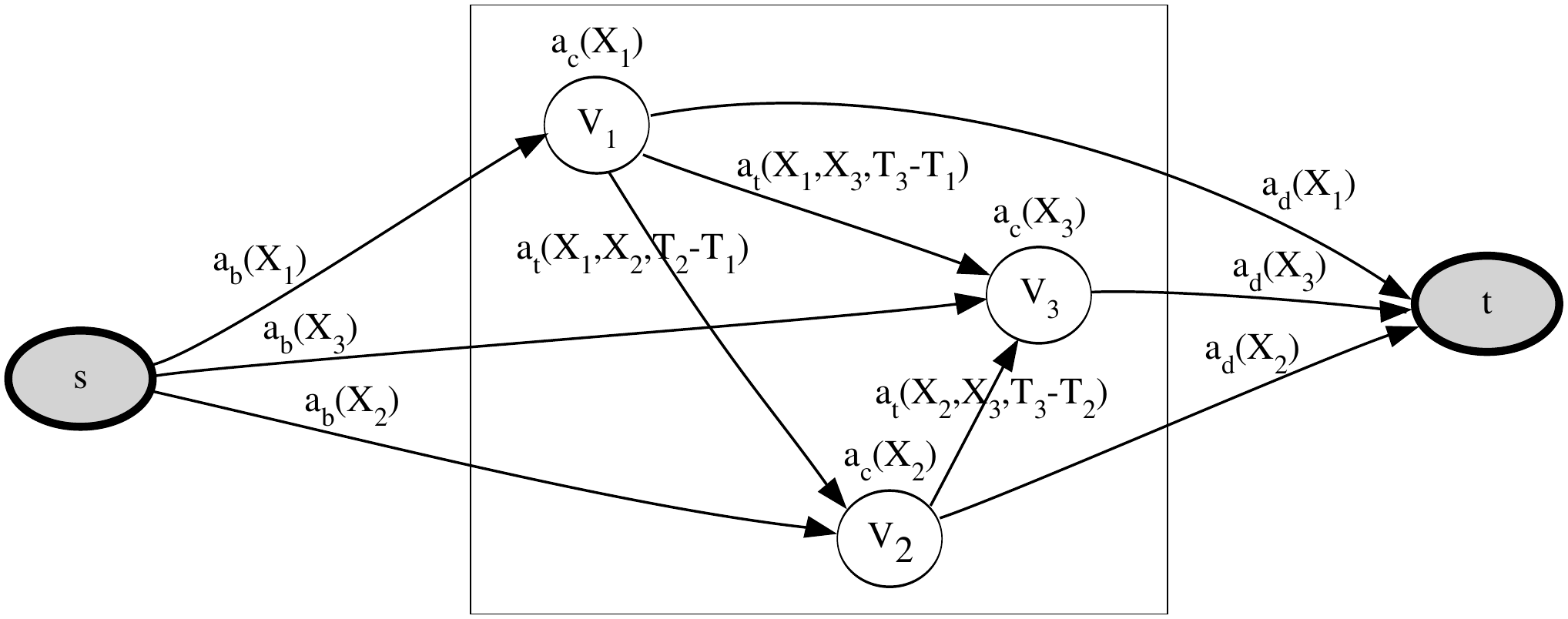}
	\caption{Constructing the weighted flow network for a set of three observations.}
\label{fig:s12t}
\end{figure*}

\begin{figure*}[t]
	\centering
	\includegraphics [width=0.7\textwidth]  {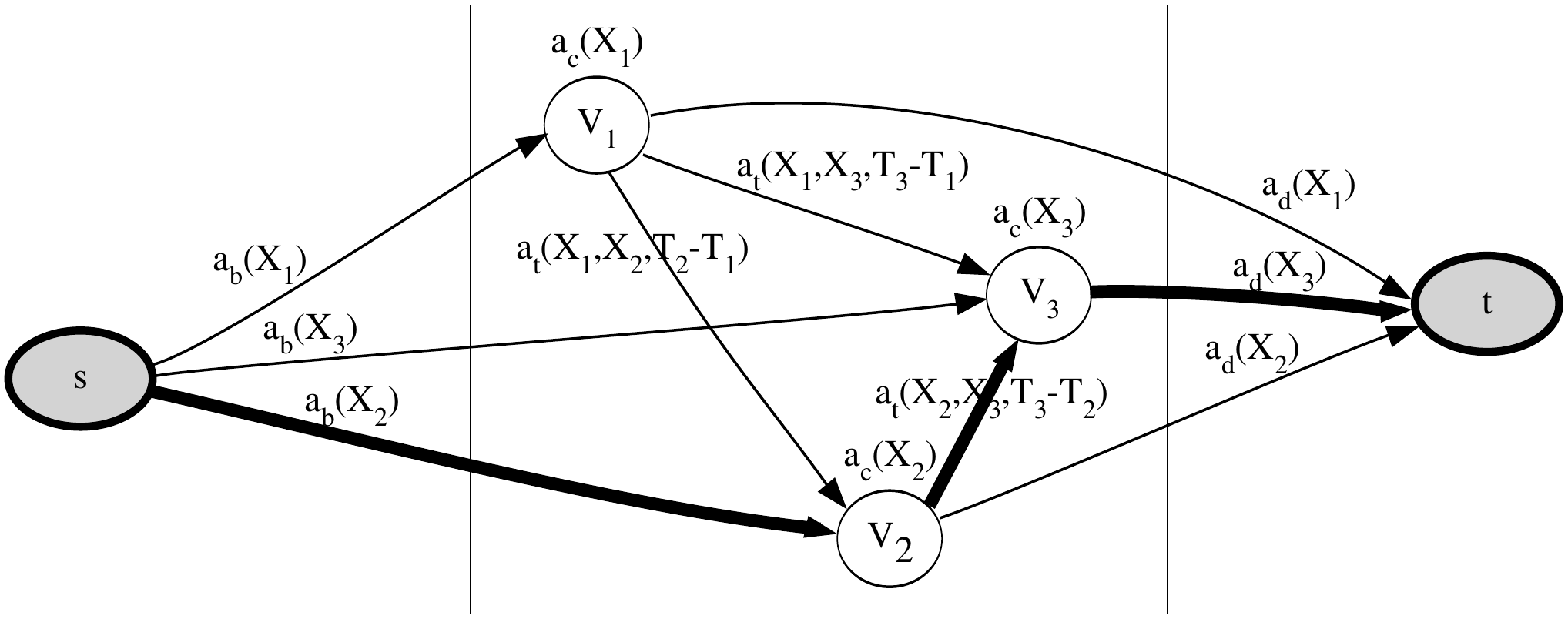}
	\caption{The network of Figure \ref{fig:s12t}, with a single-path flow indicated (\textit{s-2-3-t}).}
\label{fig:s12tflow23}
\end{figure*}

Putting these considerations together,
constructing the directed graph % $D = (V, A, a)$ 
proceeds as follows:

\begin{itemize}
	\item	A unit-capacity vertex $V_i$ is created corresponding to each observation $(X_i, T_i)$.
		The clutter noise cost $a_c(X_i)$ is associated with this vertex.
	\item	A unit-capacity arc $A_{ij}$ is created corresponding to each possible transition between two observations such that $T_i < T_j$.
		The transition cost $a_t(X_i, X_j, T_j - T_i)$ is associated with this arc.
	\item	A ``source'' vertex $s$ is added, with one arc $A_{si}$ leading from $s$ to each of the observation vertices.
		The birth cost $a_b(X_i)$ is associated with each arc $A_{si}$.
	\item	A ``sink'' vertex $t$ is added, with one arc $A_{it}$ leading from each of the observation vertices to $t$.
		The death cost $a_d(X_i)$ is associated with each arc $A_{it}$.
\end{itemize}
The temporal ordering of observations means that the graph will contain no cycles.

An illustration of the network constructed for a set of three observations is given in Figure \ref{fig:s12t}.
It is clear that 
any path from the source $s$ to a sink $t$ (we call this an ($s,t$)-path)
visits a sequence of vertices representing a temporal sequence of observations.
In the case given in Figure \ref{fig:s12t}, seven different ($s,t$)-paths are possible, and various combinations of these can form a feasible flow.
For example the flow along the single path \textit{s-2-3-t} highlighted in Figure \ref{fig:s12tflow23}
represents the possibility that the observations $X_2$ and $X_3$ were generated by a single MRP while $X_1$ is clutter:
the costs associated with flow along that path (the \textit{path flow}) are related to the birth of \textit{2}, the transition from \textit{2} to \textit{3}, and the death of \textit{3},
plus the clutter noise costs.
The cost associated with any single-path flow corresponds to one of the $K$ top-level summands in Equation \eqref{eqn:logor}.
Since in our case each (s,t)-path carries one unit of flow, 
the \textit{value} of each feasible flow is the number of paths it contains,
and corresponds to the number of MRP processes inferred in the data.
The total \textit{cost} of each feasible flow is the sum of the path costs contained,
and corresponds to the sum calculated in Equation \eqref{eqn:logor}.

\subsection{Inference}
\label{sec:inference}

The minimum cost flow in a network constructed according to our scheme
corresponds to the clustering with maximum likelihood ratio.
So to perform inference we can use existing algorithms that solve minimum-cost network flow problems.
%\citep[Chapter 3]{Bang-Jensen:2009}.
The \textit{value} of the minimum-cost flow, which gives the number of MRP sources inferred,
may be any integer between $0$ and $N$.
We use the Edmonds-Karp algorithm
\citep[Chapter 3]{Bang-Jensen:2009},
which iteratively searches for single paths in a \textit{residual network} representation
and does not get trapped in local optima.
The Edmonds-Karp algorithm
is often used to find maximum-value flow but can be used to optimise cost in our case of binary capacities.

We now consider the time complexity of our inference.
The asymptotic time complexity of the Edmonds-Karp search relates to the number of vertices and arcs as $O(|V| |A|^2)$.
The number of vertices is closely related to the number of observations $N$;
since we generate an arc for every possible transition between a pair of observations, 
$|A|$ may be on the order of $N^2$ in the worst case.
Hence we add a constraint in constructing the arcs which is reasonable in many applications:
we assert that transitions have an upper limit in the size of the time step,
and so we do not create arcs for time separations above some threshold $\tau_{\mathrm{max}}$.
The cardinality $|A|$ is then on the order of $N B$ where $B$ is the maximum number of observations within a time window of size $\tau_{\mathrm{max}}$ (and often $B << N$).

If faster search is required at the cost of optimality, greedy search strategies are available.
One such strategy is to repeatedly apply a minimum-cost path algorithm to the network,
at each iteration taking the resulting path as an identified cluster and removing its vertices from the network before the next iteration.
Since the graph is acyclic, finding a minimum-cost path can be performed very efficiently with order $O(|A|+|V|)$ at each iteration
\citep[Section 2.3.2]{Bang-Jensen:2009};
however there is no guarantee of optimality
since the overall minimum-cost flow is not guaranteed to be composed of path flows of lowest individual cost.
In our experiments we will compare this greedy search empirically against the optimal search.

In the present work we primarily consider offline (batch) inference.
However, online inference is possible within the same framework, 
in which new observations are received incrementally
by updating the graph as observations arrive.
The Edmonds-Karp search cannot be used on such a dynamic network,
except by re-starting the search from scratch upon update.
Alternative strategies such as those based on cycle-cancelling can be used to provide an updateable inference
\citep[Section 3.10.1]{Bang-Jensen:2009}.
The speed of cycle-cancelling relative to Edmonds-Karp may depend on the nature of the data;
we implemented both and found the cycle-cancelling relatively slow.

Thus far we have considered inference using a single set of MMRP model parameters,
encoded as the costs in \eqref{eqn:logor}.
It may be of value to evaluate the same data under different MMRP models,
in situations where multiple types of MRP process (having different parameters) may be active.
Multiple parametrisations cannot be represented together in a single flow network since they
would assign conflicting costs to arcs.
To accommodate incompatible costs is equivalent to 
the ``multi-commodity'' extension of the minimum-cost flow problem,
which is NP-complete \citep{Even:1975}.
However, if the clutter noise model is held constant between two different MMRP inferences,
then the two likelihood ratios calculated by \eqref{eqn:or} can be divided through to
give a likelihood ratio between the two.
This allows us to choose between possible MMRP models although not to combine them in a single clustering.

To summarise the MMRP inference described in this section:
given a set of observations plus MRP process parameters and noise process parameters,
one first represents the data as a flow network, with added source and sink nodes, and with costs representing component likelihoods (Section \ref{sec:flow}).
One then applies a minimum-cost flow algorithm to the network such as Edmonds-Karp.
Each ($s,t$)-path in the resulting minimum-cost flow represents a single cluster (a single MRP sequence) in the maximum-likelihood result,
while the nodes which receive no flow represent data to be labelled as noise.

%%%%%%%%%%%%%%%%%%%%%%%%%

\providecommand{\fsn}{F_\text{SN}}
\providecommand{\ftrans}{F_\text{trans}}

%%%%%%%%%%%%%%%%%%%%%%%%%
\section{Experiments}
\label{sec:experiments}

We have described a multiple Markov renewal process (MMRP) inference technique which takes an MRP model, an iid clutter noise model and a set of timestamped data points,
and finds a maximum-likelihood partition of the data into zero or more MRP sequences plus clutter noise.
In the following, we will illustrate its properties with a synthetic experiment (Section \ref{sec:multitest}),
before applying it to a specific task of tracking multiple singing birds in an audio mixture (Section \ref{sec:chch}).
We must first consider how to evaluate algorithm outputs.

\subsection{Evaluation Measures}
\label{sec:fstats}

To judge the empirical performance of our inference procedure, we must determine whether it can correctly separate signal from noise,
and whether it can correctly separate each individual MRP sequence into its own stream.
MMRP inference can be considered as a clustering task and could be evaluated accordingly.
However, the noise cluster is qualitatively different from the MRP clusters,
and the transitions within MRP sequences are the latent features of primary interest,
so we will focus our evaluation measures on signal/noise separation and transitions.

In the following our
statistics will be based on the standard F-measure \citep[Chapter 5]{Witten:2005},
which summarises precision and recall as follows:
\begin{align}
	F &= 2 \cdot \frac{\text{precision} \cdot \text{recall}}{\text{precision} + \text{recall}}  
	\label{eqn:ftrans1}
	\\
	&= \frac {2 t_+ }{(2 t_+ + f_- + f_+)}
	\label{eqn:ftrans2}
\end{align}

\noindent
where $t_+$ is the number of true positive detections, 
$f_+$ the number of false positive detections (noise data labelled as signal),
and $f_-$ the number of false negative detections (signal data labelled as noise).
However, the task for which our MMRP inference is designed is not an ordinary classification task:
the signal/noise label for each ground-truth datum can be treated as a class label to be inferred,
but the individual signal streams to be recovered do not have labels. %:
To quantify performance we use the F-measure in two ways.
The first (which we denote $\fsn$) evaluates the signal/noise classification performance
without considering the clustering.
The second (which we denote $\ftrans$) evaluates the performance at recovering the \textit{pairwise 
transitions} that are found in the ground-truth signals,
i.e.\ the arcs in the true dependency graph underlying the data.

\begin{figure}[t]
	\centering
	\includegraphics [width=0.4\textwidth]  {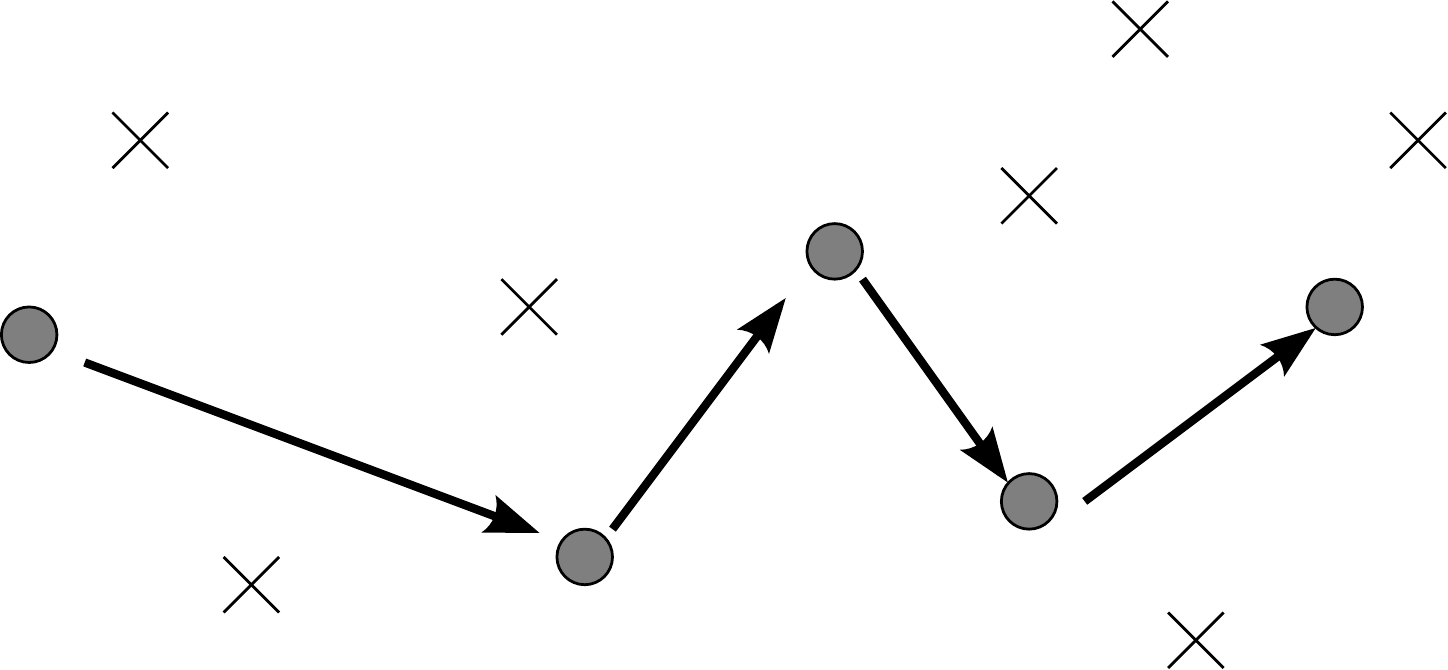} \\
	\bigskip 
	\includegraphics [width=0.4\textwidth]  {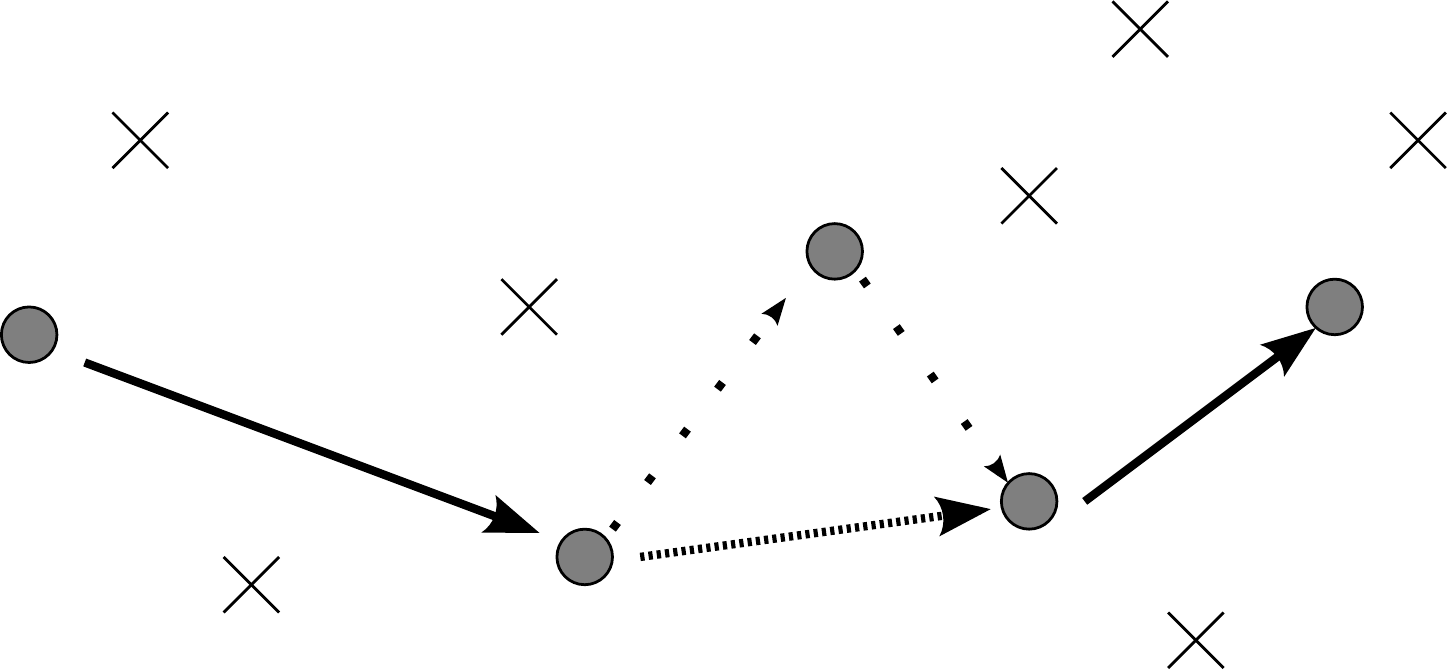}
	\caption{Illustration of errors reflected in $\ftrans$. %
The upper diagram shows a hypothetical ground-truth transition through a sequence of five observations (circles) accompanied by clutter noise (crosses). %
The lower diagram shows what would happen if inference missed one of those observations out of the chain, %
resulting in two false-negatives (dotted arrows) for ground-truth transitions not recovered, %
plus one false-positive (dashed arrow) for a transition that does not exist in the ground-truth. %
Considering these as well as the two true-positives and applying \eqref{eqn:ftrans2}, the $\ftrans$ value here is $\frac{4}{7}$.
	}
\label{fig:ftrans}
\end{figure}

To illustrate $\ftrans$, consider a situation in which
a ground-truth sequence was perfectly recovered except that one datum in the middle was left out
(Figure \ref{fig:ftrans}).
This would correspond to a number of true positives, but also two false negatives (the omission of the transition into and out of the missing datum)
and one false positive (the mistaken inference of a transition from the missing datum's predecessor to its follower).

Correctly-classified noise observations do not affect $\ftrans$ since they are not associated with any signal transitions.
Thus, $\fsn$ is useful to measure signal/noise separation while $\ftrans$ provides complementary information about correctly recovering separate streams.

\subsection{Synthetic Experiment}
\label{sec:multitest}

For our synthetic experiment we generated data in a one-dimensional state space,
with dependency structures inspired by the classic ``audio streaming'' experiments used to explore human auditory grouping of sound sequences \citep{Winkler:2012}.

\begin{figure}[t]
	\centering
	\includegraphics [width=0.8\textwidth, clip, trim=15mm 75mm 11mm 0mm]  {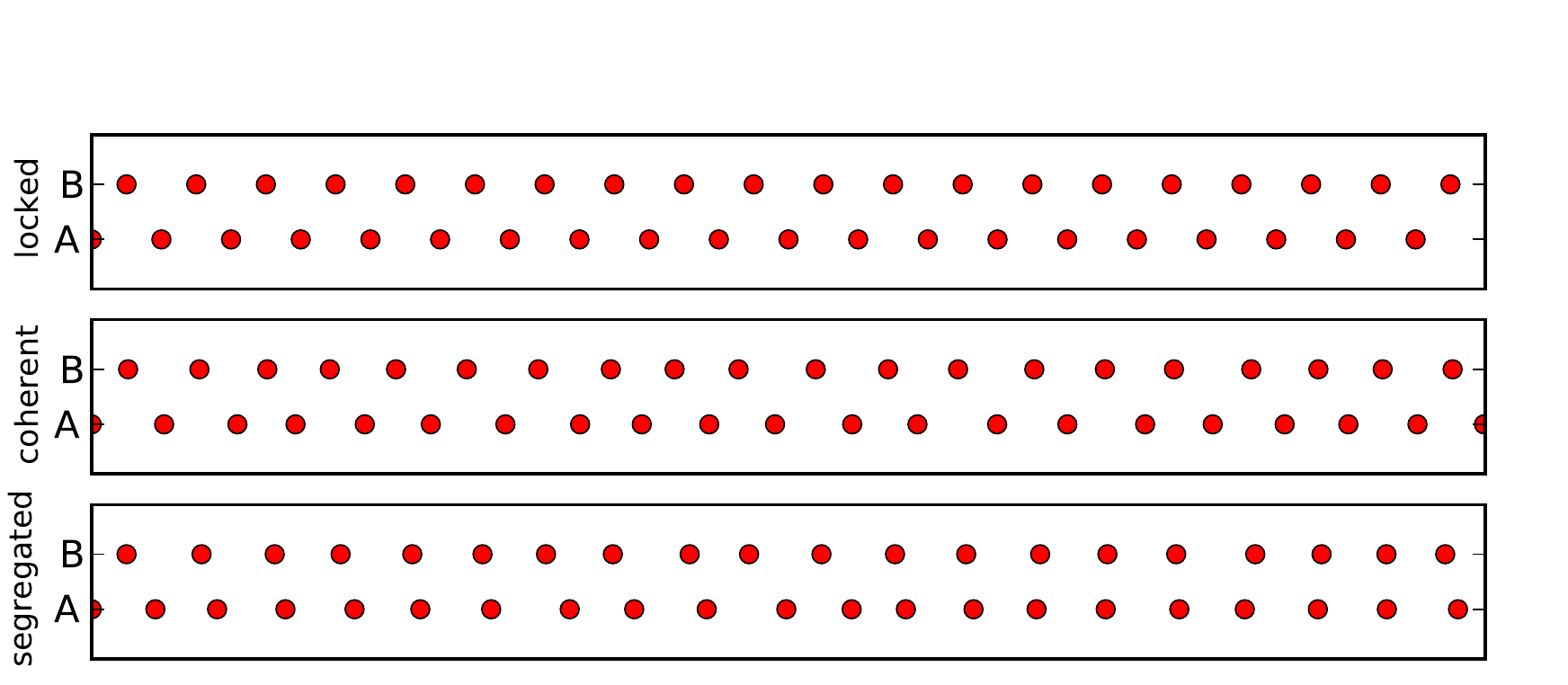} \\
	\includegraphics [width=0.8\textwidth, clip, trim=15mm 45mm 11mm 96mm]  {output_pdf/plot_generators}
	\caption{Examples of sequences generated by strict locked ABABAB repetition (top), and by similar generators but with time offsets affected by process noise reflecting either coherent (ABABAB, middle) or segregated (A\_A\_A\_ and \_B\_B\_B, bottom) dependency structure.}
\label{fig:plotgenerators}
\end{figure}

A strictly alternating sequence of the form ABABAB\dots, where A and B are different tones (Figure \ref{fig:plotgenerators}, top row),
can be interpreted either as a single alternating sequence (the ``coherent'' interpretation) 
or as a simultaneous but out-of-phase pair of constant sequences (the ``segregated'' interpretation).
Various factors can lead an observer to prefer one interpretation or the other;
here we focus on the case where drift in the timing of the events makes one or the other model more likely \citep[Experiment 2]{Cusack:2000}.
If the sequences drift such that the phase of the As and Bs remain in constant relationship (Figure \ref{fig:plotgenerators}, second row),
this is consistent with a ``coherent'' alternating generator, though may by chance be generated by a ``segregated'' pair of generators.
If the sequences drift such that the phase relationship is not maintained (third row), then this is inconsistent with the ``coherent'' model
but consistent with the ``segregated'' model.
We can generate data with these properties and observe how the MMRP inference behaves under the assumptions of each model.

\begin{figure}[t]
	\centering
	\includegraphics [width=0.3\textheight]  {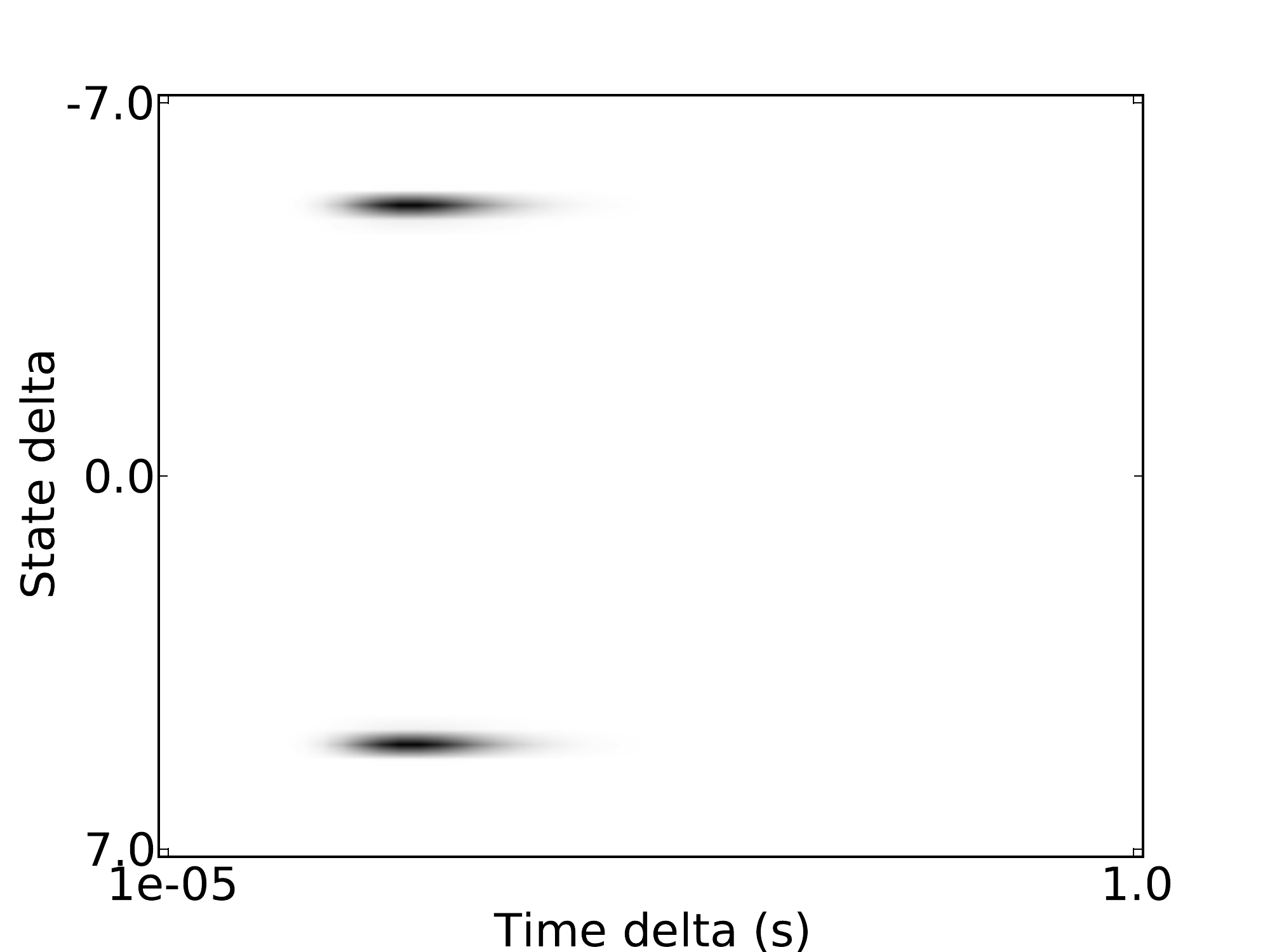}
	\includegraphics [width=0.3\textheight]  {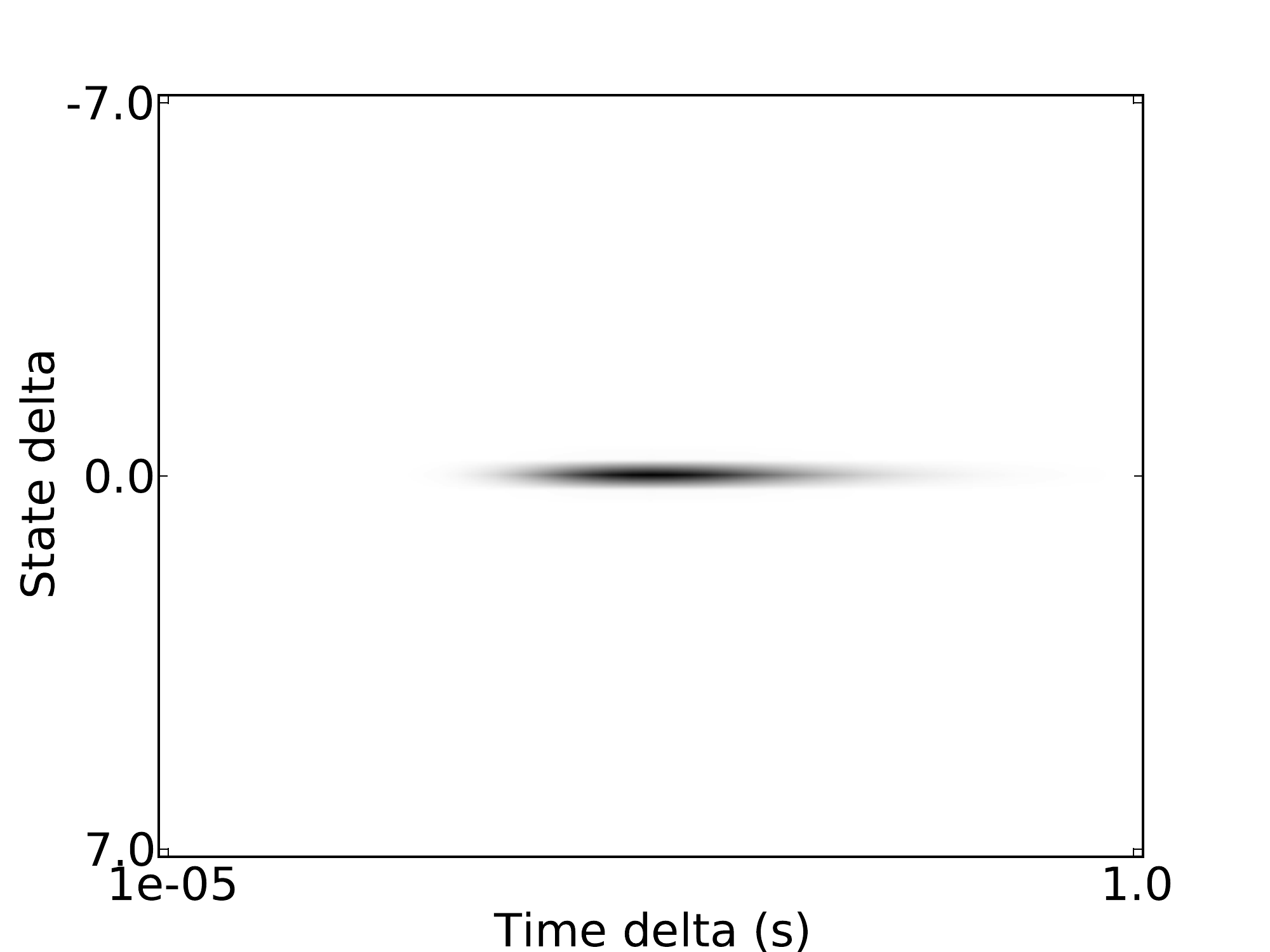}
	\caption{MRP transition probability densities for the two synthetic models: coherent (upper) and segregated (lower).}
\label{fig:plotcohseg}
\end{figure}

For our synthetic experiment we defined two separate MRP transition models (one ``coherent'' and one ``segregated'') to emit values in a one-dimensional state space $\mathcal{X} \in \mathbb{R}$.
Each model was specified by a Gaussian mixture probability distribution defined on state-delta and log-time-delta:
\begin{align}
   P(\tau_{n+1} \le t, X_{n+1}=j|X_n=i) \nonumber \\
   = f(X_{n+1}-X_n, \log{\tau_{n+1}})
\end{align}
Figure \ref{fig:plotcohseg} illustrates the transition models.
Time differences here are modelled as log-Gaussian to reflect a simple yet perceptually plausible model for lower-bounded time intervals.
The variance of the Gaussian components leads to process noise, and the two models tend to output different sequences in general.
We also define a ``locked'' model for generation only, 
which generates a strict ABABAB sequence with no process noise.
Its emissions could in principle be explained by either of the two other models.

These models served two roles in our experiment, to synthesise data and to analyse it.
For synthesis, we generated one, two or four simultaneous sequences each with a random offset in state space,
and we also added iid Poisson clutter noise in the same region of state space,
whose intensity is held constant within each run to create a given SNR.
In the case of the segregated model, each generator was a pair of such models, independent except for the initial phase and offset,
generating As and Bs as was done in Figure \ref{fig:plotcohseg}.

\begin{figure*}[t]
	\centering
	\includegraphics [width=0.9\textwidth]  {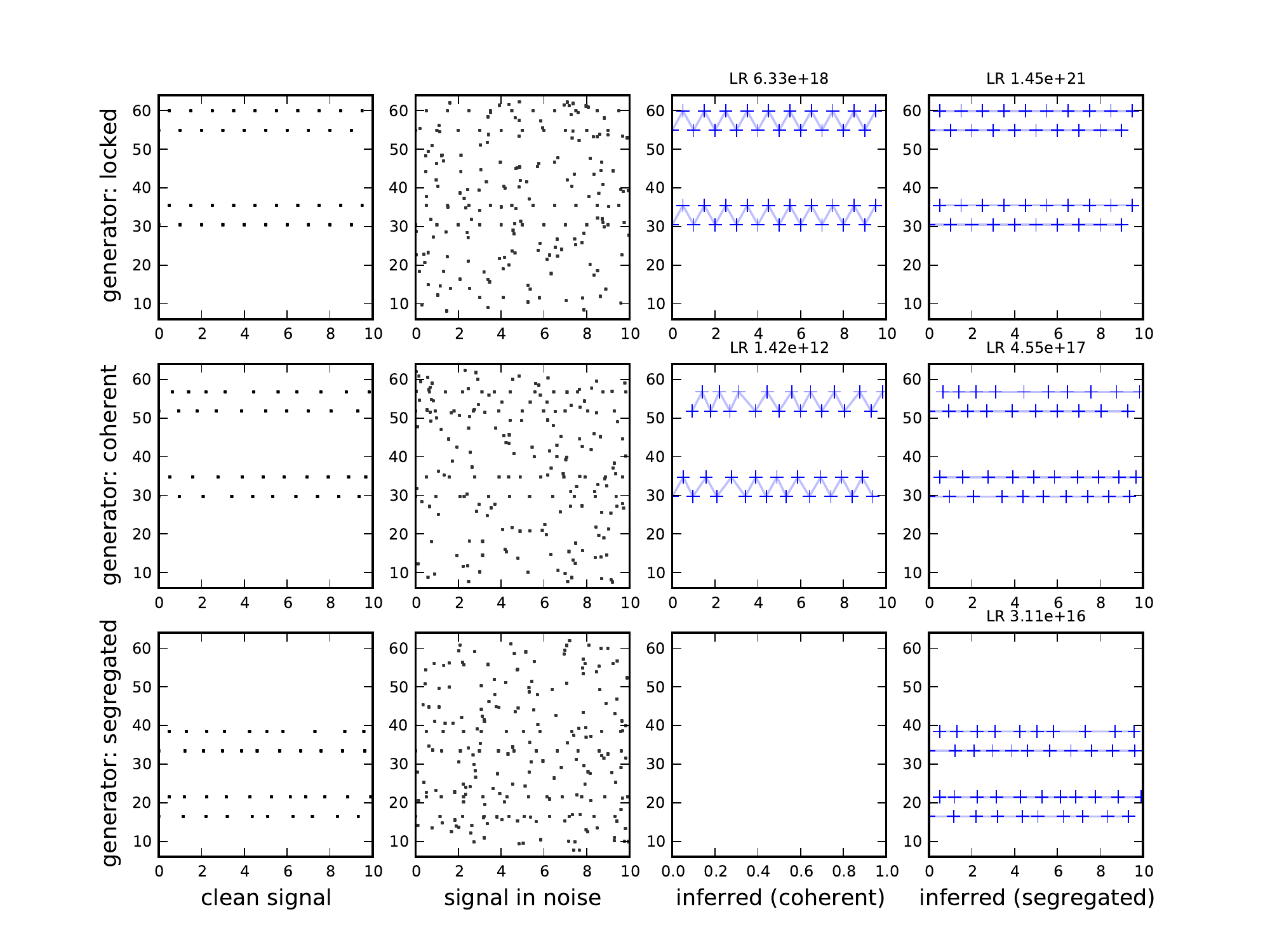}
	\caption{Results of generating observations under the locked, coherent or segregated model (in each row), and then analysing them using the coherent model or the segregated model (final two columns).}
\label{fig:plotabagenxlinlog}
\end{figure*}

The first column of Figure \ref{fig:plotabagenxlinlog} shows the results of generating data under the locked, coherent and segregated models,
with two generated sequences present in each case.
The second column shows the sequences with added clutter noise at an SNR of -12 dB.
The final two columns show the maximum-likelihood signal sequences inferred under the coherent and the segregated model.
The MMRP inference typically extracts clear traces corresponding to the ground-truth signals, even in strongly adverse SNR.
It is visually evident in the first column that the generated sequences in the middle row have some drift in their rate,
but stay in order, while the As and Bs in the bottom row drift relative to each other and do not maintain order.
This leads to unlikely emission sequences as judged by the coherent model,
and so the coherent model finds the maximum-likelihood solution to be that with no sequences (the blank plot in the figure).
Inference using the segregated model extracts traces in all three cases,
since the phase-locked drift of the coherent model is not unlikely under the segregated model.

\begin{figure}[t]
	\centering
	\includegraphics [height=0.3\textwidth, clip, trim=7mm 4mm 18mm 11mm]  {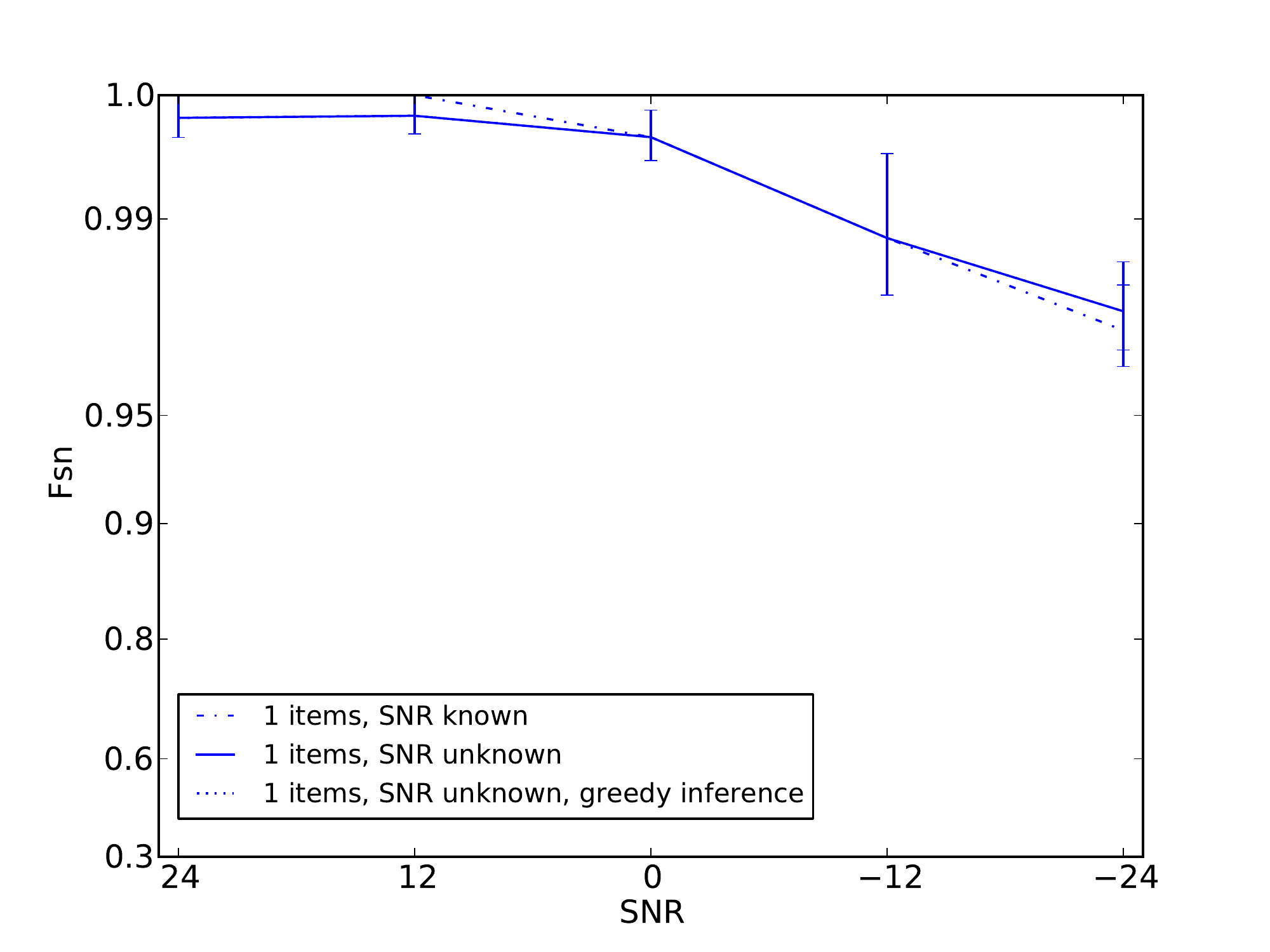}
	\includegraphics [height=0.3\textwidth, clip, trim=25mm 4mm 18mm 11mm]  {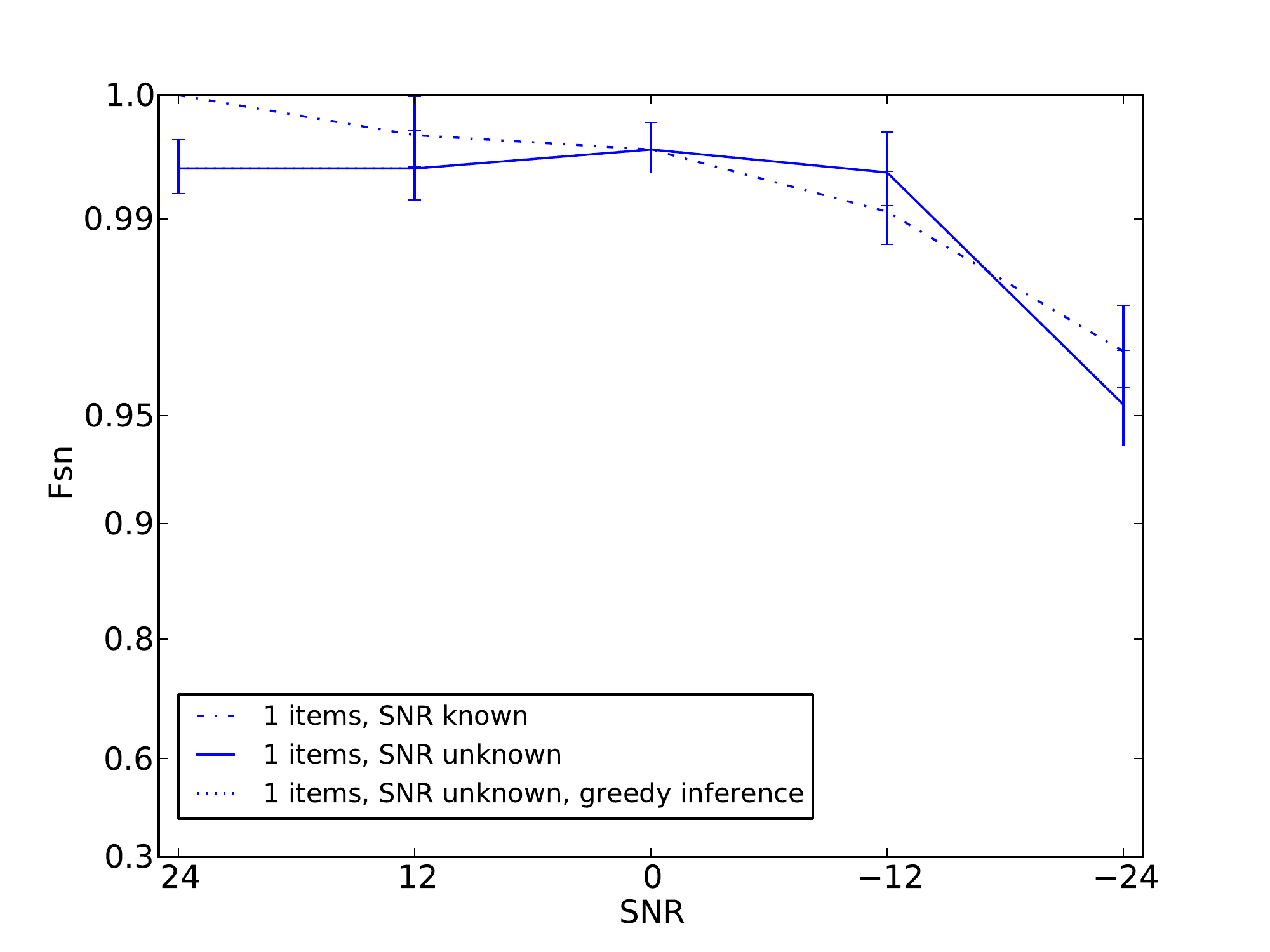} 
	\\
	\includegraphics [height=0.3\textwidth, clip, trim=7mm 4mm 18mm 11mm]  {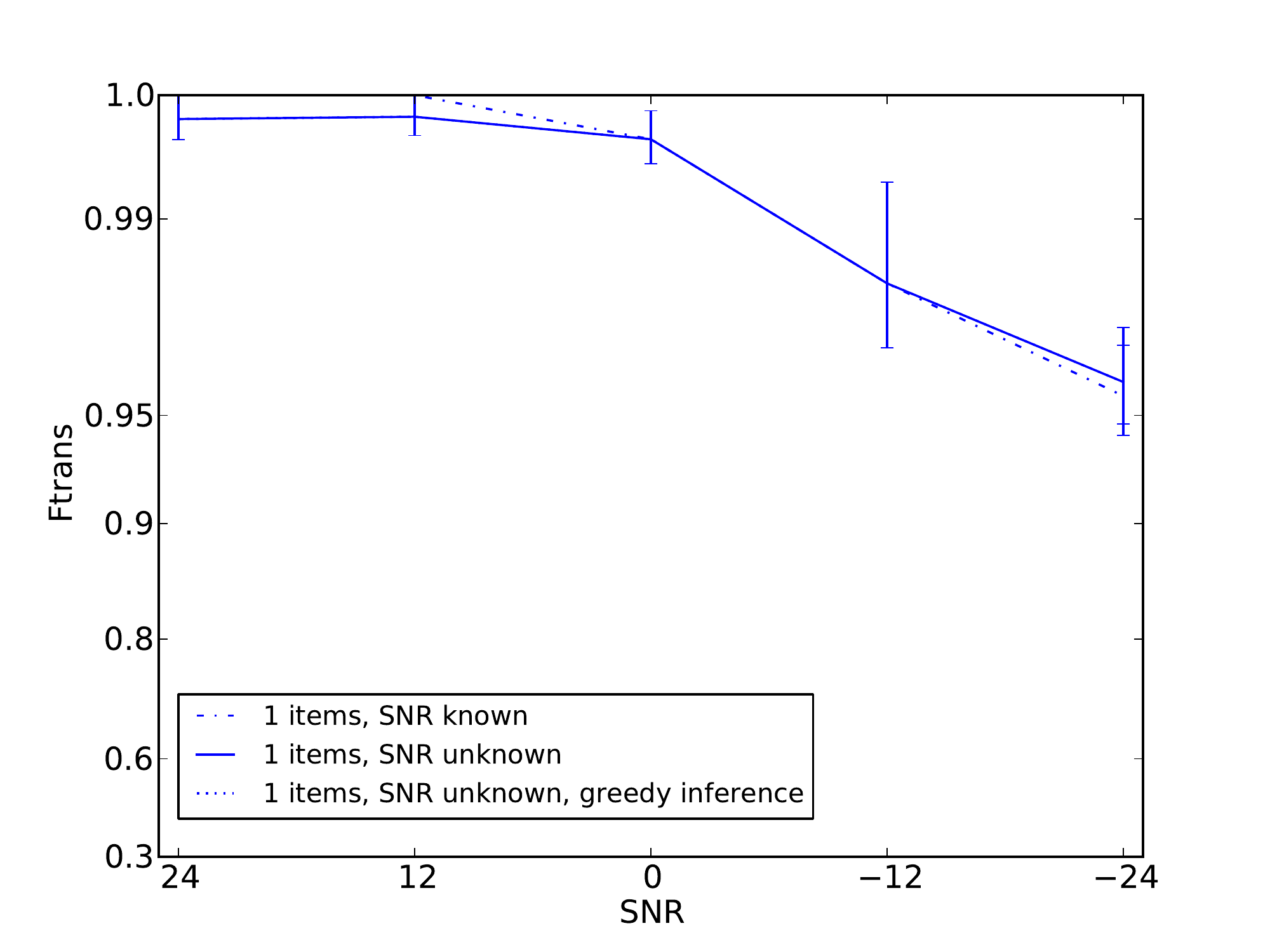}
	\includegraphics [height=0.3\textwidth, clip, trim=25mm 4mm 18mm 11mm]  {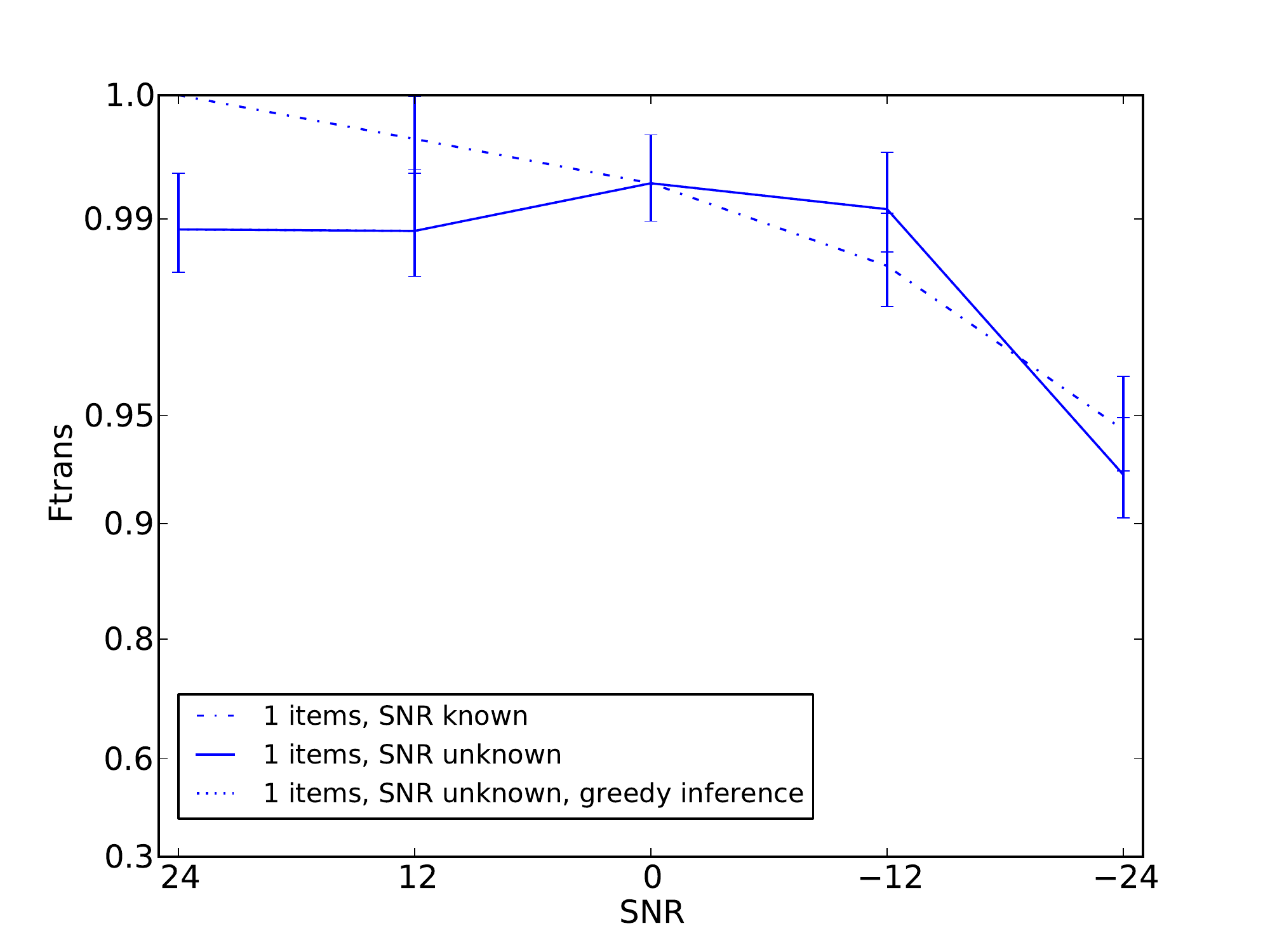}
	\caption{F-measure for signal/noise separation ($\fsn$, upper) and transitions ($\ftrans$, lower).
	The ground truth in each case is a single ABABAB stream, generated via the coherent (left) and segregated (right) cases. %
	Means and standard errors are shown; the vertical axis is reverse-log-scaled so that the results very near 1.0 can be distinguished.}
\label{fig:plotmultitest1}
\end{figure}

\begin{figure}[t]
	\centering
	\includegraphics [height=0.3\textwidth, clip, trim=7mm 4mm 18mm 11mm]  {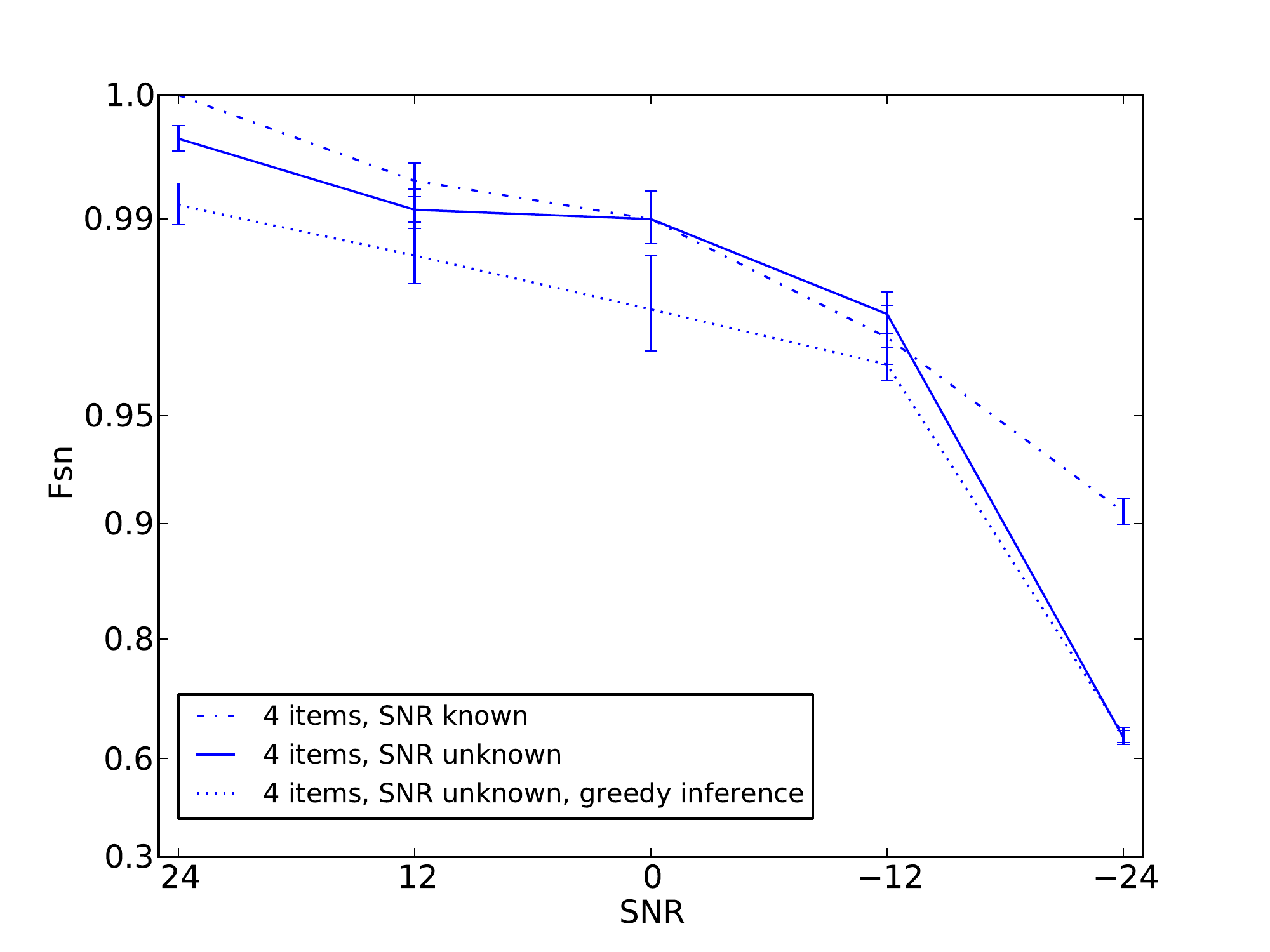}
	\includegraphics [height=0.3\textwidth, clip, trim=25mm 4mm 18mm 11mm]  {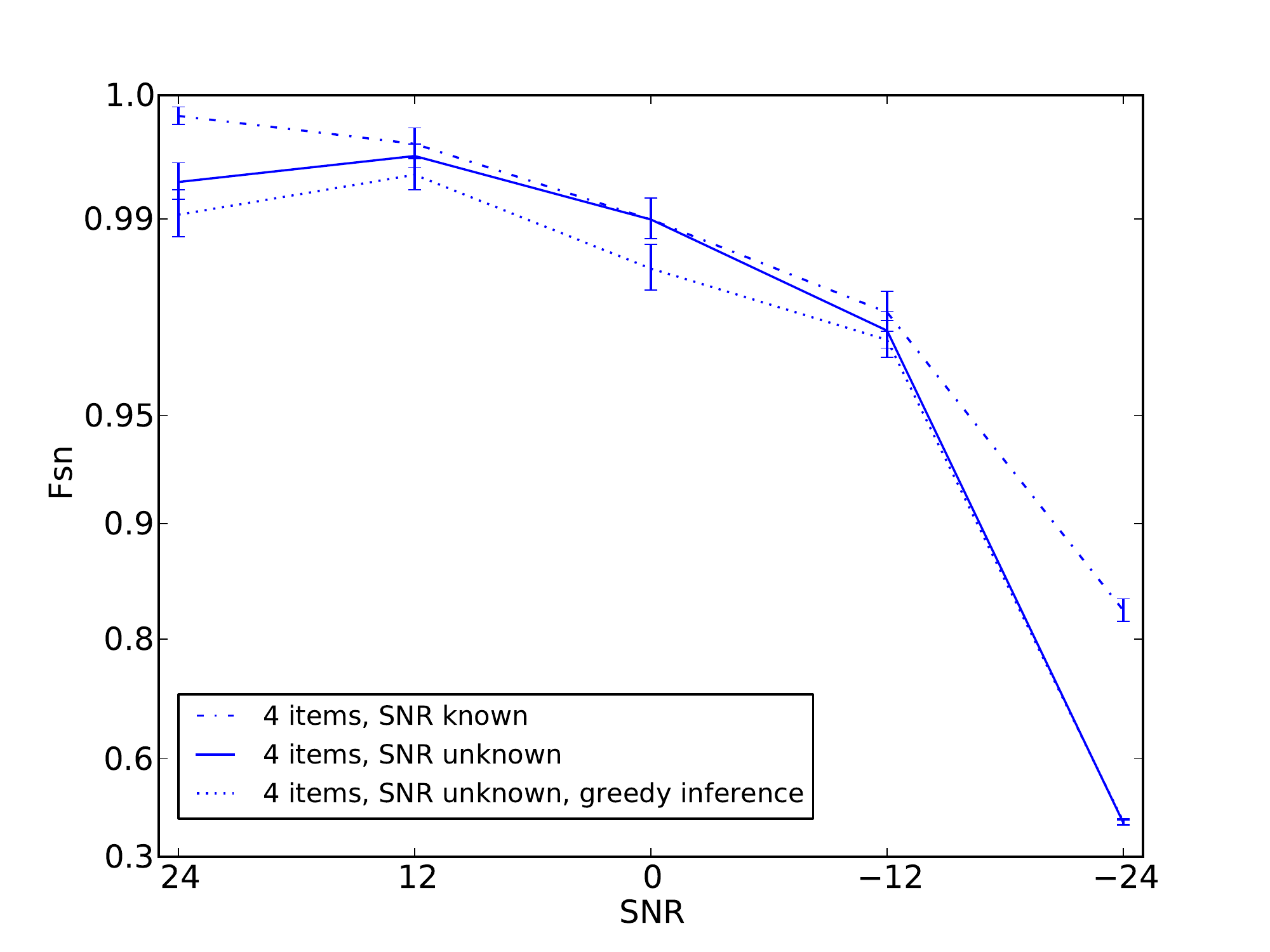} 
	\\
	\includegraphics [height=0.3\textwidth, clip, trim=7mm 4mm 18mm 11mm]  {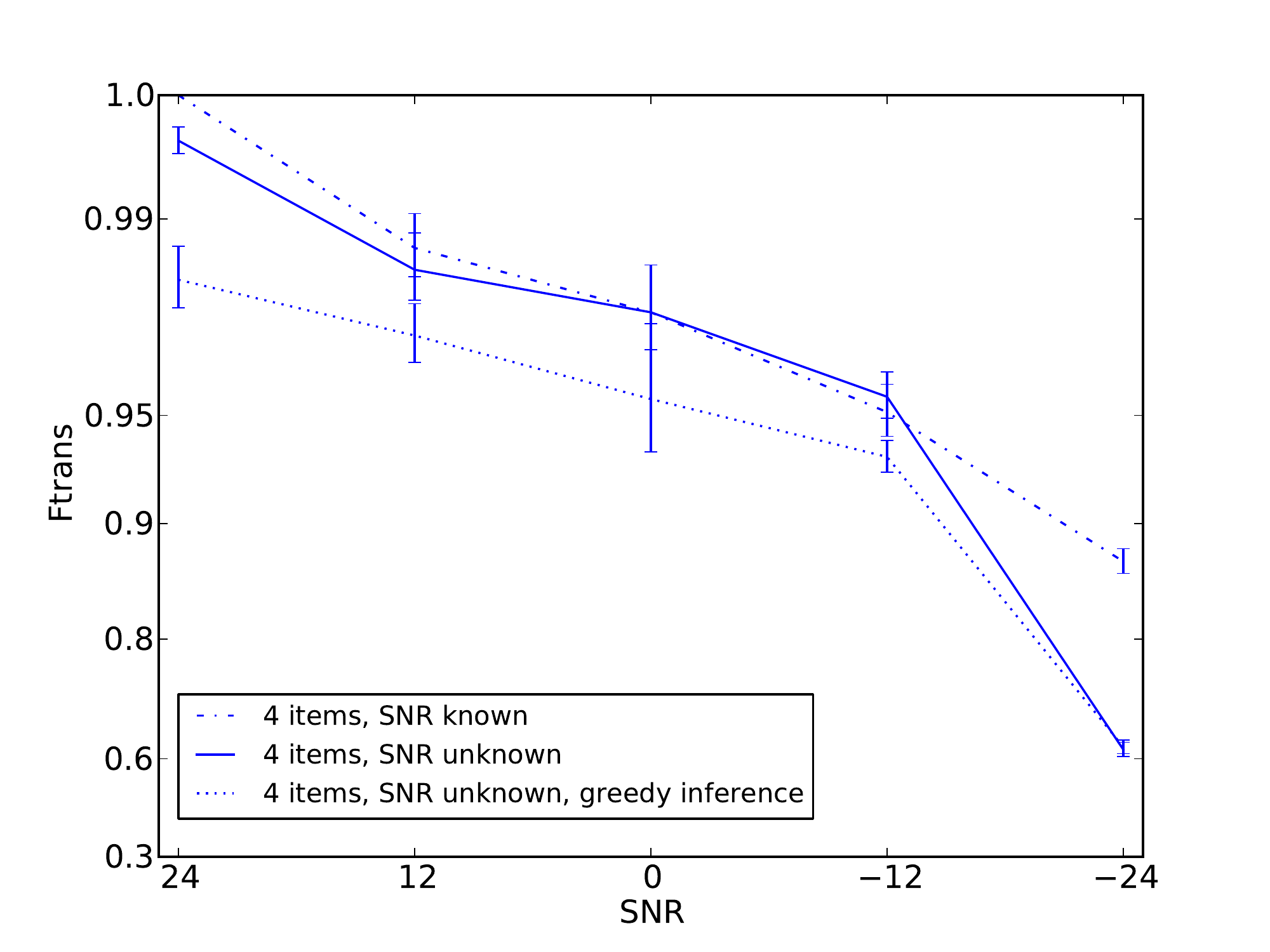}
	\includegraphics [height=0.3\textwidth, clip, trim=25mm 4mm 18mm 11mm]  {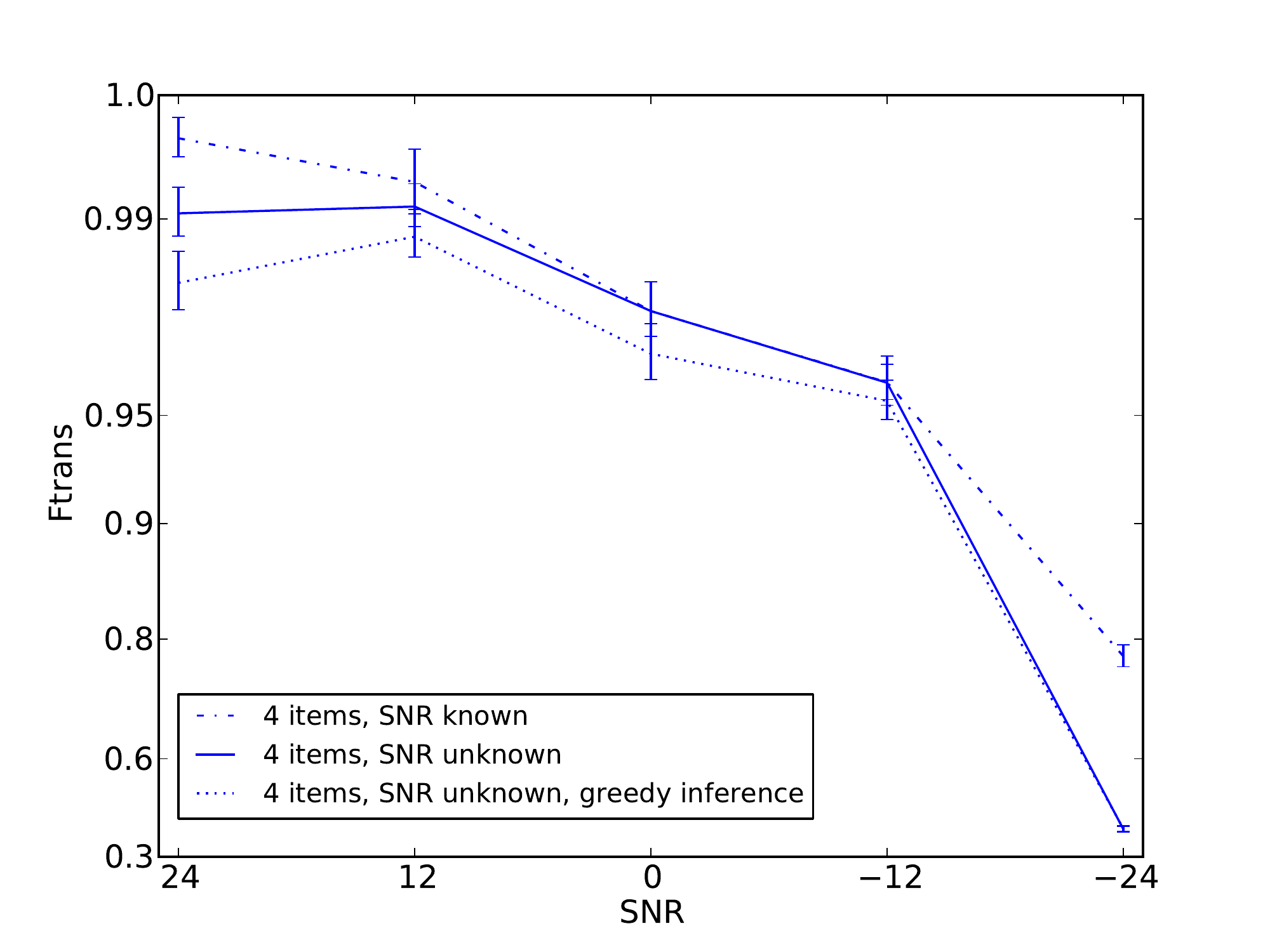}
	\caption{As Figure \ref{fig:plotmultitest1} but with four simultaneous generated streams rather than one.}
\label{fig:plotmultitest4}
\end{figure}

To evaluate our inference procedure, we ran this process 
multiple times,
varying the SNR level, the number of items present, and whether the true SNR was known to the algorithm.
When not known, the SNR estimate was arbitrarily held fixed at 0 dB.
We tested both the optimal and greedy inference algorithms described in Section \ref{sec:inference}.
For each setting we conducted 20 runs and recorded the $\fsn$ and $\ftrans$ statistics.
Figures \ref{fig:plotmultitest1} and \ref{fig:plotmultitest4}
illustrate the results, and show a consistent pattern according to both statistics.
Recovery performance is very strong in all but the most adverse conditions, in most cases being well above 0.95.
For these particular scenarios, recovery is impaired under the strongest condition tested (4 simultaneous generators and SNR -24 dB).
Under other conditions the recovery is good, whether the true SNR is known to the algorithm or not.
Knowing the true SNR does not add a clear improvement to performance, showing that the inference is robust to the SNR estimate parameter.
Greedy inference has lower time complexity than the full inference, 
but 
when there are multiple streams to be recovered
it yields poorer performance than the full algorithm 
even at very favourable SNR.


\subsection{Birdsong Audio Experiment}
\label{sec:chch}

Many natural sound sources produce signals with structured patterns of emissions and silence, for example birdsong or footsteps.
If the emissions due to one such source can be modelled as an MRP, 
then our inference procedure should be able to separate multiple simultaneous ``streams'' of emissions.
In the following experiment we studied the ability of our inference to perform this separation in data derived from audio signals containing multiple instances of a species of bird common in many European countries,
the Common Chiffchaff \citep{Salomon:1992}.
Chiffchaff song consists of sequences of typical length 8--20 ``syllables''.
Each syllable is a pitched note consisting of a downward chirp to a briefly-held tone in the region of 5--8 kHz.
Syllables are separated by around 0.2--0.3 seconds.
The exact note sequence has not to our knowledge been studied in detail;
it appears to exhibit only short-range dependency,
and is thus amenable to analysis under Markovian assumptions.

\subsubsection{Data Preparation}
\label{sec:chchdata}

To aid reproducibility, we used recordings from the Xeno Canto database of publicly-available bird recordings.%
\footnote{\url{http://www.xeno-canto.org/europe}} %
We located 25 recordings of song of the Chiffchaff (species \textit{Phylloscopus collybita})
recorded in Europe (excluding any recordings marked as having ``deviant'' song or uncertain species identity; also excluding \textit{calls} which are different from \textit{song} in sound and function).
The recordings used are listed in Table \ref{tbl:chchdataset}.
We converted the recordings to 44.1 kHz mono wave files, 
high-pass filtered them at 2 kHz, and normalised the amplitude of each file.

\begin{table}[p]
\centering
\begin{tabular}{l l}
        ID & Country \\
        \hline
XC103404 & pl \\
XC25760 & dn \\
XC26762 & se \\
XC28027 & de \\
XC29706 & se \\
XC31881 & nl \\
XC32011 & nl \\
XC32094 & no \\
XC35097 & es \\
XC35974 & cz \\
XC36603 & cz \\
XC36902 & nl \\
XC46524 & nl \\
\end{tabular}
\begin{tabular}{l l}
        ID & Country \\
        \hline
XC48263 & no \\
XC48383 & de \\
XC54052 & it \\
XC55168 & fr \\
XC56298 & de \\
XC56410 & ru \\
XC57168 & fr \\
XC65140 & es \\
XC77394 & dk \\
XC77442 & se \\
XC97737 & uk \\
XC99469 & pl \\
\\
\end{tabular}
\caption{Chiffchaff audio samples used in our dataset, giving the Xeno Canto ID and the country code. %
 Each recording can be accessed via a URL such as \texttt{http://www.xeno-canto.org/XC103404}, %
 and the dataset is also archived at \texttt{http://archive.org/details/chiffchaff25} %
}
\label{tbl:chchdataset}
\end{table}

Each audio file was analysed separately to create training data;
during testing, audio files were digitally mixed in groups of two to five files.

\begin{figure}[t]
	\centering
	\includegraphics [width=0.25\textwidth]  {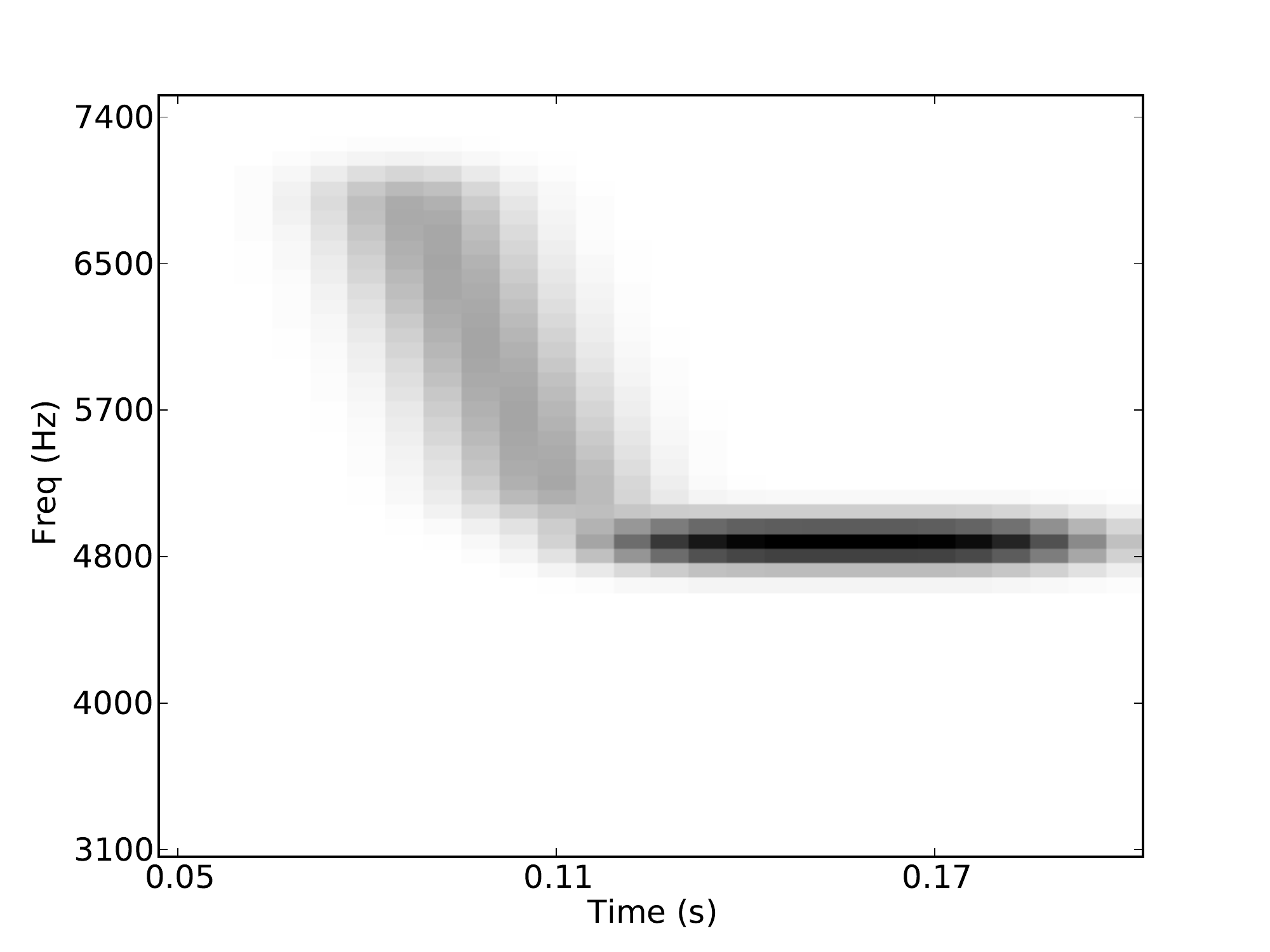}
	\caption{Template used for spectro-temporal cross-correlation detection. The downward and horizontal bars have equal total weight; the latter appears darker because shorter. The template is a manually-constructed Gaussian mixture model having 40 components.}
\label{fig:plotxcorgrid}
\end{figure}

In order to convert an audio file into a sequence of events amenable to MMRP inference,
we used spectro-temporal cross-correlation to detect individual syllables of song, as used by \citet{Osiejuk:2000}.
We designed a spectrotemporal template using a Gaussian mixture (GM) to represent the main characteristics of a single Chiffchaff syllable,
a downward chirp to a briefly-held note
(Figure \ref{fig:plotxcorgrid}).
The GM was modelled on a Chiffchaff recording from Xeno Canto which was not included in our main dataset (ID number XC48101).
Then to analyse an audio file we converted the file into a spectrogram representation (512 samples per frame, 50\% overlap between frames, Hann window),
and converted the GM to a sampled grid template with the same time-frequency granularity as the spectrogram,
before sliding the grid template along the time axis and along the frequency axis (between 3--8 kHz),
evaluating the correlation between the template and spectrogram at each location.
Correlation values were treated as detections if they were local peaks with value greater than a threshold correlation of 0.8.

\begin{figure}[t]
	\centering
	%\hfill
	\includegraphics [width=0.46\textwidth, clip]  {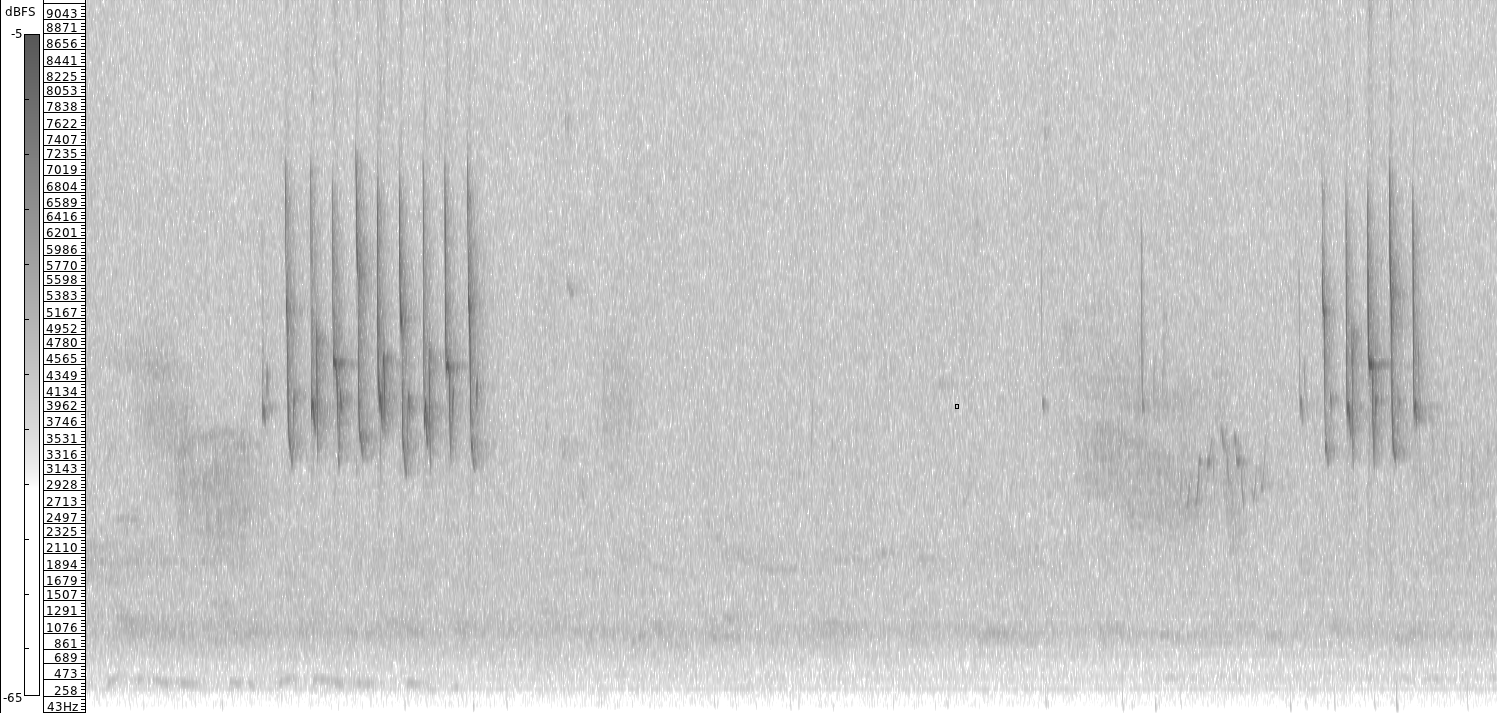}
	\\
	%\hfill
	\includegraphics [width=0.5\textwidth, clip, trim=8mm 5mm 16mm 14mm ]  {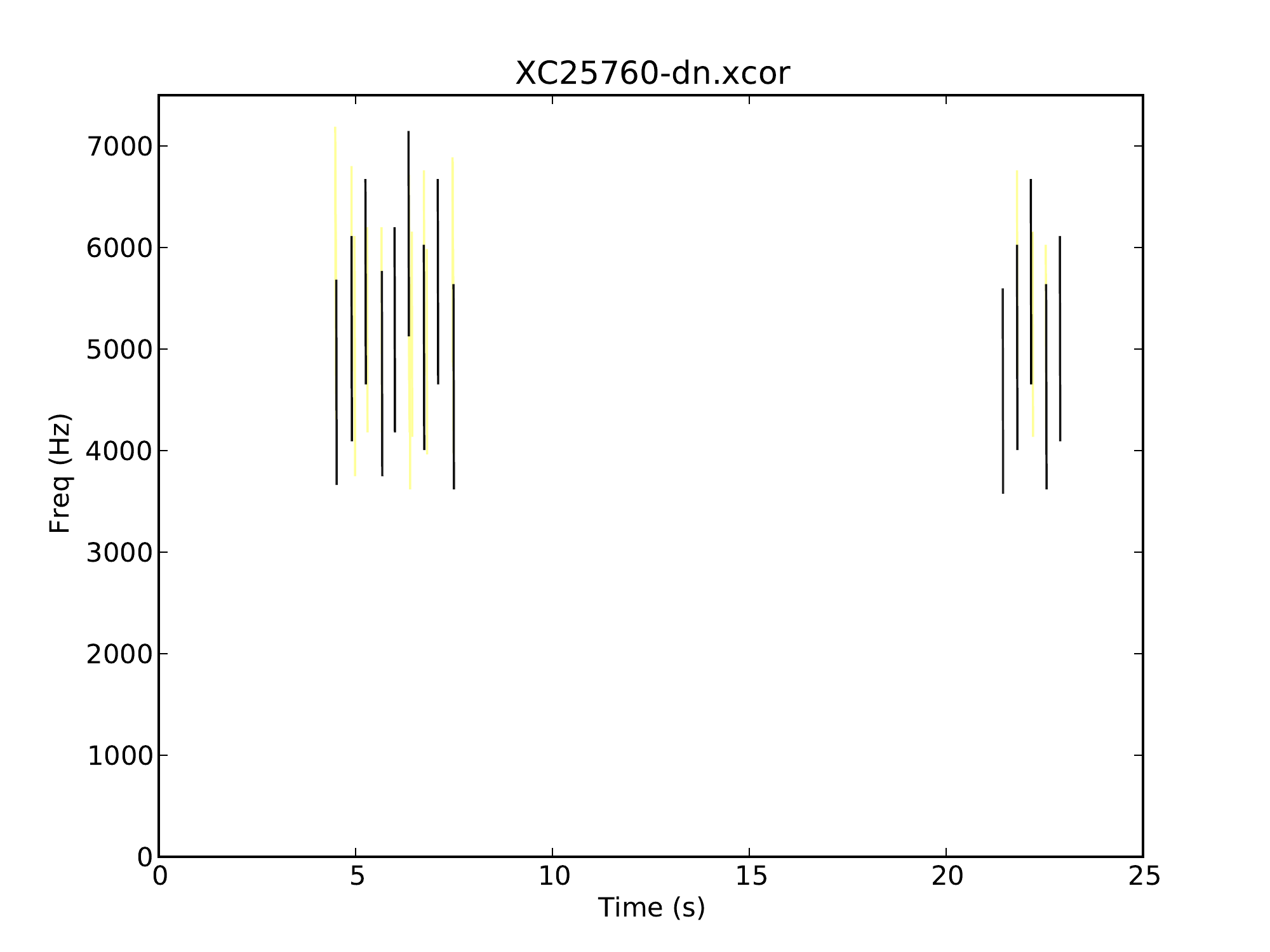}
	\caption{Example of cross-correlation detection: excerpt of spectrogram shown (top), and the corresponding detections (bottom). %
	In the lower image, bold lines represent detections treated as ``signal'' in the filtering used for training, while the fainter lines represent detections used to train the noise model. %
	}
\label{fig:plotxcorexample}
\end{figure}

Such cross-correlation detection applied to an audio file produces a set of observations, each having a time and frequency offset and a correlation strength (Figure \ref{fig:plotxcorexample}).
It typically contains one detection for every Chiffchaff syllable,
with occasional doubled detections and spurious noise detections.
When applied to mixtures of audio, this produces data appropriate for MMRP inference.

In order to derive a Gaussian mixture model (GMM) transition probability model from monophonic Chiffchaff training data,
for each audio file in a training set we filtered the observations automatically to keep only the single strongest detection within any 0.2 second window.
This time limit corresponds to the lower limit on the rate of song syllables;
such filtering is only appropriate for monophonic training sequences and was not applied to the audio mixtures used for testing.
The filtered sequences were then %assumed to be free of noise detections and doubled detections, and
used to train a 10-component GMM with full covariance,
defined on the vector space having the following four dimensions:
\begin{itemize}
	\item	log(frequency) of syllable one
	\item	log(frequency) of syllable two
	\item	log(magnitude ratio between syllables) 
	\item	log(time separation between syllables)
\end{itemize}

We also trained a separate GMM to create a noise model, taking the set of observations that had been discarded in the above filtering step
and training a 10-component GMM with full covariance to fit an iid distribution to the one-dimensional log(frequency) data for the noise observations.

\subsubsection{Inference from Audio Mixtures}

In order to test whether the MMRP approach could recover syllable sequences from audio mixtures,
we performed an experiment using five-fold cross-validation.
For each fold we used 20 audio files for training,
and then with the remaining five audio files we created audio mixtures of up to five signals,
testing recovery in each case.

The quality of signal/noise separation and of clustering the syllables correctly could depend on various features of the experimental task,
including 
whether observations could be extracted from audio mixtures as reliably as from single recordings,
the generalisability of the fitted GMMs, 
noise levels,
and the MMRP inference procedure.
In order to explore these factors we compared various different analysis approaches:

\begin{description}
\item[Audio recovery:]
	The primary approach was to take a mixture audio file, apply spectro-temporal cross-correlation as described above,
	then to apply MMRP inference using the signal and noise GMMs.
\item[Audio recovery (greedy):]
	This approach was as above, but using greedy recovery rather than the optimal flow inference.
\item[Ideal recovery:]
	There is no guarantee that the same observations will be recovered from the mixture audio as were recovered from the individual recordings.
	To simulate ideal-case recovery, instead of using the audio mixture we simply pooled the signal and noise observations that had been derived from
	the test set's individual mono analysis, then performed MMRP inference as in the audio recovery case.
\item[Ideal recovery, synthetic noise:]
	To simulate ideal recovery but with more adverse noise conditions, we proceeded as in the ideal case, but also added extra clutter noise at 0 dB.
	To do this, we created a copy of every observation in the test set, but assigned it an independent random time position, thus creating noise with the same 
	frequency distribution as the true signal.
\item[Ideal recovery, tested on training set:]
	To measure an ``upper limit'' on performance and probe the generalisation capability of the algorithm, we proceeded as in the ideal case, but used GMMs trained on the actual test files to be analysed
	rather than on the separate training data. If this resulted in stronger performance than the ideal-case, it would indicate issues with generalising to signals outside the training set.
\item[Audio recovery, baseline:]
	In order to provide a low-complexity baseline showing the recovery quality using only the marginal properties of the signal and noise, we created a simple
	baseline system which treated both signal and noise as iid one-dimensional log(frequency) data,
	using maximum likelihood to label each observation as either signal or noise.
	The baseline system then clustered together observations that were identified as signal and were separated by less than 0.7 seconds.
\end{description}

We tested each of these approaches using mixtures of one, two, three, four or five of the test recordings.
As in the previous experiment, we measured the $\fsn$ statistic to evaluate signal/noise separation,
and the $\ftrans$ statistic to evaluate the performance at recovering separate sequences.

\begin{figure*}[t]
	\centering
	\includegraphics [width=0.5\textwidth, clip, trim=7mm 3mm 18mm 11mm]  {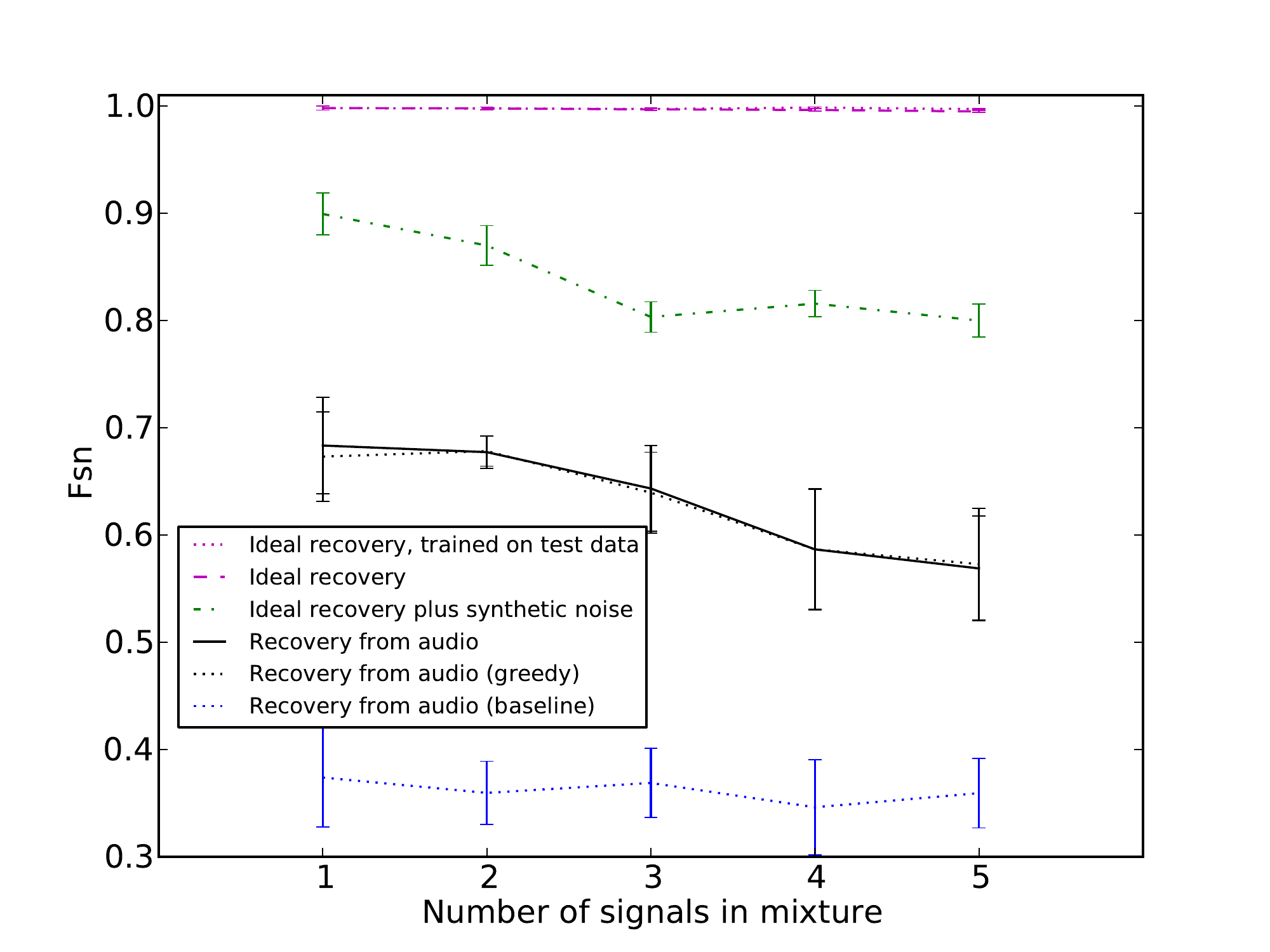}
	\includegraphics [width=0.5\textwidth, clip, trim=7mm 3mm 18mm 11mm]  {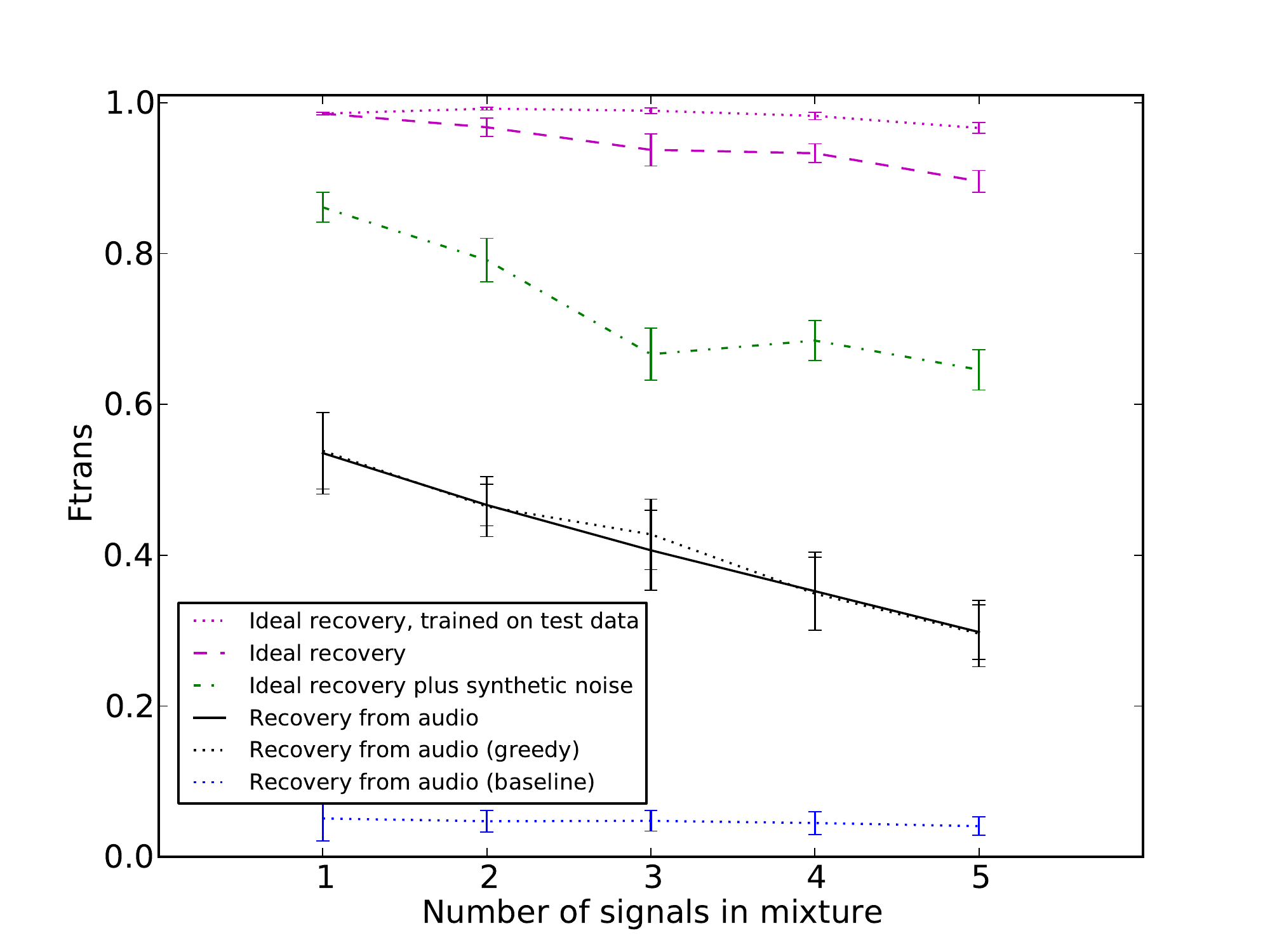}
	\caption{The $\fsn$ and $\ftrans$ %
evaluation measures for the Chiffchaff audio analyses. %
Means and standard errors are shown taken over the five folds of the cross-validation.}
\label{fig:plotchchstats}
\end{figure*}

Results are shown in Figure \ref{fig:plotchchstats}.
Although the two 
statistics we measure reflect different aspects of performance,
they both
rank the different analysis approaches in a very similar way.
All the MMRP inference runs exhibit a significant and very strong improvement over the baseline.
Very strong performance is achieved in the noiseless ``ideal recovery'' cases,
achieving results similar to those in the previous synthetic experiment.
The small size of the difference between training on the test data and on the training data
indicates that the algorithm can generalise across the data used in our experiment.

When synthetic noise is added to the ideal-recovery case, performance is reduced by a moderate but consistent amount.
When we use recovery from audio mixtures, performance reduces again.
This shows that the practical task of retrieving detections from audio mixtures has a significant effect on the algorithm performance.
However, even in this case our algorithm outperforms the baseline system by a very wide margin,
showing the value of MMRP inference for separating signal from noise and clustering signals into MRP streams.

As we increase the number of recordings in the mixture, performance of all the analysis approaches shows a mild decline.
However even with five recordings the performance of the MMRP remains relatively strong.
%Encouragingly, performance of all the analysis approaches remains stable as we increase the mixture from two to five simultaneous streams.

In this experiment, unlike the previous one,
we see very little difference between the performance of the full inference and the greedy inference.
Thus the faster greedy inference is appropriate in some but not all situations;
in this experiment it is not a limiting factor in performance.

%%%%%%%%%%%%%%%%%%%%%%%%%
\section{Conclusions}
\label{sec:conc}

In this paper we have introduced a specific clustering problem, that of segregating time\-stamped data originating in multiple point processes plus clutter noise.
We developed an approach to inferring structure in data produced by a mixture of
an unknown number of similar Markov renewal processes (MRPs) plus independent clutter noise.
The inference simultaneously distinguishes signal from noise as well as clustering signal observations
into separate source streams,
by solving a network flow problem isomorphic to the MMRP mixture problem.

In a synthetic experiment we have shown that inference can perform very well even under high noise conditions 
(up to $-24$ dB SNR).
In an experiment on birdsong audio data we have also shown strong performance,
albeit with a dependence on the quality of the underlying representation to recover events from audio data.
Our method is general and has very few free parameters. %, and is available as open-source Python code.%
%\footnote{TODO LATER: publish code online}

The inference in the present work is limited to models without hidden state and with only single-order Markov dependencies.
These limitations arise from the combinatorial ambiguity in MMRP mixtures (unlike ordinary Markov models)
over which is the immediate predecessor for each observation.
Future work will aim to find techniques to broaden the class of models that can be treated in this way.

%%%%%%%%%%%%%%%%%%%%%%%%
\section*{Acknowledgments}
\label{sec:ack}

\textit{(Acknowledgments to be added in final version.)}
% ACK TODO: funding sources, ALSO ADD A LICENCE TO THE PAPER.

%%%%%%%%%%%%%%%%%%%%%%%%%
%\end{document}

%%%%%%%%%%%%%%%%%%%%%%%%%
\bibliographystyle{plainnat}
\bibliography{../refs}

\end{document}